\documentclass[10pt,letterpaper]{article}

\usepackage{cogsci}

\cogscifinalcopy 

\usepackage{pslatex}
\usepackage{apacite}
\usepackage{float} 

\usepackage{graphicx}
\usepackage{multirow}
\usepackage{booktabs}
\usepackage{amsfonts}       
\usepackage{amsmath}        
\usepackage{algorithm}      
\usepackage{algpseudocode}  
\usepackage{listings}       
\usepackage{enumitem}       
\usepackage{xcolor}         
\usepackage{subcaption}     

\newcommand*{\block}[1]{%
  \framebox{\raisebox{0pt}[0.4\baselineskip][0.00\baselineskip]{#1}}
}

\title{Infinite Ends from Finite Samples:
Open-Ended Goal Inference as \\ Top-Down Bayesian Filtering of Bottom-Up Proposals}
 
\author{
  {\large \bf Tan Zhi-Xuan, Gloria Kang, Vikash Mansinghka, Joshua B. Tenenbaum} \\
  Department of Brain and Cognitive Sciences, MIT \\
  Correspondence to \texttt{xuan@mit.edu}
  }

\begin{document}

\maketitle

\begin{abstract}
The space of human goals is tremendously vast; and yet, from just a few moments of watching a scene or reading a story, we seem to spontaneously infer a range of plausible motivations for the people and characters involved. What explains this remarkable capacity for intuiting other agents' goals, despite the infinitude of ends they might pursue? And how does this cohere with our understanding of other people as approximately rational agents? In this paper, we introduce a sequential Monte Carlo model of \emph{open-ended goal inference}, which combines top-down Bayesian inverse planning with bottom-up sampling based on the statistics of co-occurring subgoals. By proposing goal hypotheses related to the subgoals achieved by an agent, our model rapidly generates plausible goals without exhaustive search, then filters out goals that would be irrational given the actions taken so far. We validate this model in a goal inference task called Block Words, where participants try to guess the word that someone is stacking out of lettered blocks. In comparison to both heuristic bottom-up guessing and exact Bayesian inference over hundreds of goals, our model better predicts the mean, variance, efficiency, and resource rationality of human goal inferences, achieving similar accuracy to the exact model at a fraction of the cognitive cost, while also explaining garden-path effects that arise from misleading bottom-up cues. Our experiments thus highlight the importance of uniting top-down and bottom-up models for explaining the speed, accuracy, and generality of human theory-of-mind.

\end{abstract}

\section{Introduction}

Whether one is watching a play, reading a novel, or spending time with a friend at their house, inferences about others' goals and motivations often arise spontaneously and unbidden \shortcite{moskowitz2016spontaneous}: Is the person crouching behind a tree trying to hide from, or spy on someone? Does the strange warrior who has just entered the fray of battle intend to kill the protagonist, or save them? When your friend gets up from the couch and walks to the kitchen, are they getting a snack, or making some tea? Despite the seemingly infinite space of possible goals, we have little trouble in coming up with plausible hypotheses, and then --- as the story unfolds --- filtering out those that fail to explain our observations. If the warrior defends our protagonist from a stray arrow, they are likely an ally. If your friend opens the fridge, they are probably having a snack. What computational mechanisms underlie this ability to both hypothesize and evaluate the goals that explain others' behavior, even when the set of possibilities is vast and open-ended?

While psychologists have long studied how people both \emph{generate} \shortcite{heider1944experimental,graesser1994constructing,hassin2005automatic,van2012spontaneous} and \emph{evaluate} hypotheses about the goals that other agents have \shortcite{gergely2003teleological,jara2015children,liu2017ten}, computational models of human goal inference have focused on the latter, assuming a small and \emph{fixed} set of possible goals, then modeling how people infer their relative likelihoods \shortcite{baker2009action,ullman2009help,kleiman2016coordinate,vered2016online,jara2019naive}. This leaves open how people come up with plausible goals in the first place, especially in large hypothesis spaces where enumeration over all possibilities makes inference intractable \shortcite{kwisthout2013bridging,blokpoel2013computational}. How then are people solving this seemingly intractable problem (if they do so at all)? Even though recent advances in Bayesian inverse planning have shown how modeling the plans of other agents \shortcite{zhi2020online,alanqary2021modeling} and inferring goals from static scenes \shortcite{chandra2023inferring} can be made orders of magnitude more efficient, they do not address the key challenge posed by open-ended settings: Efficiently generating plausible goal hypotheses.

In this paper, we develop an algorithmic account of open-ended goal inference, which combines top-down Bayesian inverse planning and bottom-up sampling in a sequential Monte Carlo (SMC) algorithm \shortcite{del2006sequential}. Instead of exhaustively enumerating the space of goals, our model assumes that humans are familiar with the statistics of their environments \shortcite{griffiths2006optimal}, and can rapidly generate relevant hypotheses based on contextual, data-driven cues \shortcite{schulz2012finding, phillips2019we}. In particular, we assume familiarity with the statistics of co-occurring subgoals, such that complete goals can rapidly be generated once some subgoals have been achieved. Our model then filters these goals according to the principle of rational action \shortcite{gergely2003teleological,baker2009action}, keeping those that best explain the agent's actions. We evaluate this model in Block Words, a game where observers have to guess the word that someone is stacking out of lettered blocks \shortcite{ramirez2010probabilistic,alanqary2021modeling}. Subgoals correspond to partial words, so observers can generate plausible goals by ``auto-completion''. However, this bottom-up strategy is insufficient in general --- some goals may be irrational given the actions observed so far, necessitating inverse planning.

To test the predictions of our model, we conduct an experiment where human participants play a series of rounds in Block Words. Each round is carefully designed so as to elicit various patterns of inference --- some where bottom-up guessing is sufficient, some where inverse planning is required to filter out irrational goals, and some intended to produce garden-path inferences that exact Bayesian reasoning should avoid. As alternatives to our model, we test pure bottom-up sampling, as well as an exact Bayesian baseline that performs enumerative inference over all English words that can be spelled in each round. We compare these models by measuring their similarity to human responses in terms of their mean, variance, sample efficiency, and computational cost, allowing us to determine their fidelity to human goal inference in both behavioral and algorithmic terms.

\section{Computational Model}

Building upon prior accounts of human goal inference \cite{baker2009action,zhi2020online,alanqary2021modeling}, we assume that observers perform approximately Bayesian inference over a generative model of how other agents plan and act to achieve their goals:
\begin{alignat}{2}
\textit{Goal Prior:}& \quad g \sim P(g) \label{eq:goal-prior} \\
\textit{Online Planning:}& \quad \pi_t \sim P(\pi_t | s_{t-1}, \pi_{t-1}, g) \label{eq:online-planning} \\
\textit{Action Selection:}& \quad a_t \sim P(a_t | s_{t-1}, \pi_t) \label{eq:action-selection} \\
\textit{State Transition:} & \quad s_t \sim P(s_t | s_{t-1}, a_t) \label{eq:state-transition}
\end{alignat}

Here $g$ is the agent's goal, and at each step $t$, $\pi_t$ is the agent's current plan or policy, $a_t$ is the agent's action, and $s_t$ is the state of the environment. Given a sequence of states $s_{0:T}$ and actions $a_{1:T}$, the observer's task is to infer the goal $g$ by approximating the posterior $P(g | s_{0:T}, a_{1:T})$.
Approximating this posterior presents numerous computational challenges. Among these, our focus is on the challenge posed by \emph{open-ended} settings, where the set of possible goals $g \in \mathcal{G}$ is large or potentially infinite. In this section, we first review recent advances that render goal inference over \emph{fixed} spaces algorithmically tractable, before explaining how we can extend these ideas to open-ended spaces.

\subsection{Modeling Boundedly Rational Plans and Actions}

Since computing the posterior requires simulating the plans $\pi_t$ that an agent might follow to each goal $g$, this process is also known as \emph{Bayesian inverse planning}. In general this is a difficult problem, because planning itself is a complicated and often intractable task. However, as \citeA{zhi2020online} show, this difficulty can be alleviated by treating agents as \emph{boundedly rational planners}, who spend only limited computation at each step $t$ on planning. We adopt a more recent version of this architecture \shortcite{zhixuan2024pragmatic,ying2023inferring}, modeling agents that update a \emph{policy} $\pi_t$ (i.e. a conditional plan) that defines a Boltzmann distribution over actions $a_t$ that can be taken at state $s_{t-1}$:
\begin{equation}
    P(a_t | s_{t-1}, \pi_t) = \frac{\exp\left(-\beta  \hat Q_{\pi_t}(s_{t-1}, a_t)\right)}{\sum_{a} \exp\left(-\beta \hat Q_{\pi_t}(s_{t-1}, a)\right)}
\end{equation}
Here $\hat Q_{\pi_t}(s_{t-1}, a_t)$ denotes the estimated cost of the shortest path to goal $g$ from $s_{t-1}$ that starts with action $a_t$. As such, $\pi_t$ assigns higher probability to actions along more optimal paths to the goal, with $\beta$ controlling the degree of optimality. To compute $\hat Q_{\pi_t}(s_{t-1}, a_t)$ efficiently, we use real-time heuristic search (RTHS), which updates the $\hat Q_{\pi_{t-1}}$ values computed for $\pi_{t-1}$ via a search process guided by the $Q$-values themselves \shortcite{korf1990real,barto1995learning,koenig2006real}. This process is deterministic, and limited to a computational budget of up to size $B$ (e.g. the number of search iterations). As such, given $G = |\mathcal{G}|$ possible goals, the computational cost of simulating the plans of an agent for $T$ steps (and hence exactly inferring their goals) is $O(GBT)$.

\begin{figure}[t]
    \centering
    \includegraphics[width=0.95\columnwidth]{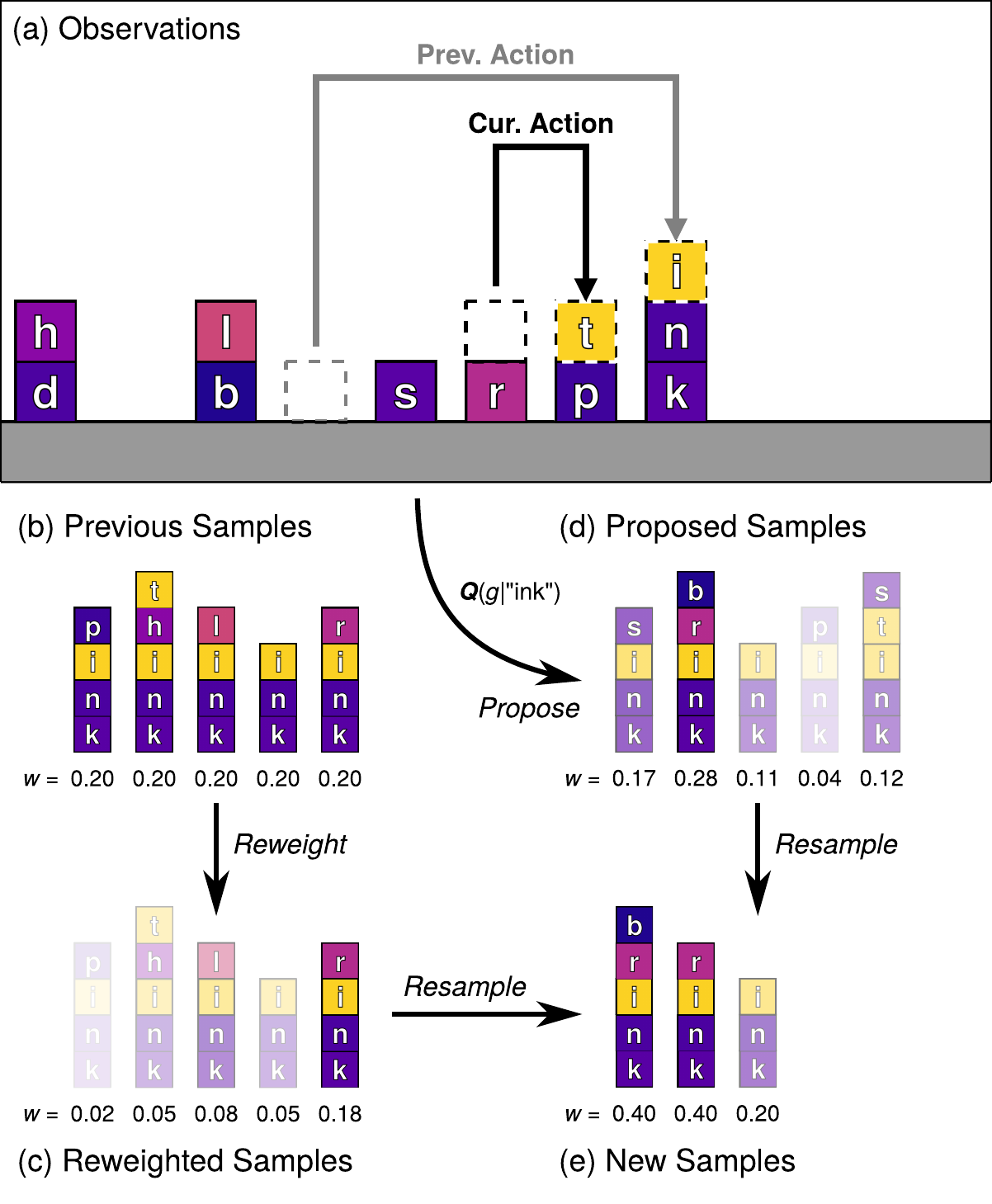}
    \caption{Illustration of open-ended goal inference in Block Words via particle filtering. Initially, \textbf{\textsf{i}} is stacked on \textbf{\textsf{n}}, leading  \textbf{\textsf{pink}} to be proposed as a goal. In the next step, however, \textbf{\textsf{t}} is stacked on \textbf{\textsf{p}}. This makes \textbf{\textsf{pink}} much less likely after reweighting, and hence removed after resampling.}
    \label{fig:example}
    \vspace{-6pt}
\end{figure}

\subsection{Bottom-Up Sampling of Plausible Goals}

While the analysis above suggests that Bayesian inverse planning is not only tractable, but \emph{linear} in computational complexity, it neglects the fact that the number of goals $G$ can grow very large. As suggested by \citeA{blokpoel2013computational}, this might be because $G$ itself grows exponentially with some other natural parameter --- in Block Words, for example, just 9-11 lettered blocks can be used to spell anywhere from 150 to 800 English words. But even without this exponential dependence, a large value of $G$ can quickly render (exact) goal inference too costly to be algorithmically plausible.

How might people manage the complexity of their inferences in these open-ended settings? We posit that in sufficiently familiar contexts, people are familiar with the statistics of co-occurring subgoals, such that given some subgoal $\gamma_i$, they can rapidly sample a complementary subgoal $\gamma_j \sim P(\gamma_j | \gamma_i)$. This means that once an agent achieves some subgoal $\gamma_i$ --- say, boiling a kettle of water, or stacking the letter \block{\textsf{n}} on top of \block{\textsf{g}} --- an observer can rapidly generate a complete goal $g = \gamma_i \land \gamma_j$ --- perhaps adding tea to the boiled water, or spelling the word \block{\textsf{s}}\block{\textsf{o}}\block{\textsf{n}}\block{\textsf{g}}. These conditional distributions can be efficiently learned using either neural networks or classical sequence models such as $n$-grams. Since the goals in our study are English words, we use a character-level $n$-gram model. Regardless of what sequence model is used, a key property is that retrieval and sampling can occur in essentially \emph{constant} time \shortcite{guthrie2010storing}, providing a plausible mechanism for relevance-guided hypothesis generation \shortcite{phillips2019we,schulz2012finding}.

\subsection{Bayesian Filtering of Bottom-Up Samples}

Given the ability to rapidly generate plausible goals, it is tempting to forgo inverse planning altogether. As Figure \ref{fig:example} illustrates, however, this strategy can go awry. Suppose you see someone stack the block \block{\textsf{i}} on top of \block{\textsf{n}}\block{\textsf{k}}, and the word \block{\textsf{p}}\block{\textsf{i}}\block{\textsf{n}}\block{\textsf{k}} comes to mind. But then you see \block{\textsf{t}} stacked on top of \block{\textsf{p}}. Is \block{\textsf{p}}\block{\textsf{i}}\block{\textsf{n}}\block{\textsf{k}} still a plausible goal? From a bottom-up perspective, \block{\textsf{p}} is still a likely completion of  \block{\textsf{i}}\block{\textsf{n}}\block{\textsf{k}}. But if we understand agents as \emph{rational} planners, this no longer seems likely. If \block{\textsf{p}}\block{\textsf{i}}\block{\textsf{n}}\block{\textsf{k}} had been the agent's goal, stacking \block{\textsf{t}} on \block{\textsf{p}} would be quite suboptimal.

If humans actually engage in the reasoning above, then modeling their inferences requires uniting top-down Bayesian inverse planning with bottom-up cues. Following other sampling-based accounts of sequential human inferences \shortcite{daw2008pigeon,vul2009explaining,thaker2017online}, we model this integration with a sequential Monte Carlo (SMC) algorithm (Algorithm \ref{alg:open-ended-sips}), extending the Sequential Inverse Plan Search (SIPS) algorithm of \shortciteA{zhi2020online}. SMC algorithms are also known as particle filters, which approximate Bayesian posteriors by maintaining a weighted set of hypotheses or particles, then updating the weights of those particles as observations arrive \shortcite{del2006sequential}. At each step, they may also \emph{resample} particles according to their weights, or \emph{rejuvenate} the particles, making perturbations to the sample collection to increase hypothesis diversity \cite{chopin2002sequential,lew2023smcp3}.

\algrenewcommand\algorithmicprocedure{\textbf{Procedure}}
\algrenewtext{EndProcedure}{\textbf{End}}
\algrenewcommand\algorithmicfor{\textbf{For}}
\algrenewtext{EndFor}{\textbf{End}}
\algrenewcommand\algorithmicreturn{\textbf{Return}}

\begin{algorithm}[t]
\caption{Open-Ended SIPS for Goal Inference}
\label{alg:open-ended-sips}
\begin{algorithmic}[1]
\footnotesize
\Procedure{Open-Ended-SIPS}{$s_{0:T}, a_{1:T}, N$}
\State \textbf{Using}: $Q(g | s_t, a_t)$, a bottom-up goal proposal.
\For{each step $t$ from $1$ to $T$}
    \State Propose $N$ new goals $g$ from $Q(g | s_t, a_t)$.
    \State Simulate policies $\pi_{1:t}$ for each new goal.
    \State Compute weights $\frac{P(g, \pi_{1:t}, s_{0:t}, a_{1:t})}{Q(g | s_t, a_t)}$ for new particles.
    \State Update policies $\pi_t$ for previous goal samples $g$.
    \State Multiply their weights by  $P(a_t | s_{t-1}, \pi_t)$.
    \State Resample full collection down to $N$ particles.
    \State Coalesce identical particles.
\EndFor
\State \Return weighted collection of $\leq N$ goal hypotheses.
\EndProcedure
\end{algorithmic}
\end{algorithm}

A variant of this rejuvenation phase is where our bottom-up samplers come in: As Algorithm \ref{alg:open-ended-sips} shows, after observing each state $s_t$ and action $a_t$ at step $t$, we use these samplers as \emph{proposal distributions} over goals $Q(g | s_t, a_t)$, generating $N$ new goal hypotheses $g$ (L4) based on bottom-up cues. These new hypotheses assigned an importance weight $P(g, \pi_{1:t}, s_{0:t}, a_{1:t}) / Q(g | s_t, a_t)$, where the numerator $P(g, \pi_{1:t}, s_{0:t}, a_{1:t})$ accounts for how well the plans $\pi_{1:t}$ that lead to $g$ explain the actions $a_{1:t}$, and the denominator $Q(g | s_t, a_t)$ compensates for $g$ having been sampled from the proposal (L5--7). Open-ended SIPS also \emph{reweights} previous samples based on how well they explain the current action $a_t$ (L7--8). Finally, we resample the particle collection back down to $N$ samples (L9), coalescing identical samples by summing their weights (L10). Our algorithm thus implements the high-level logic described earlier: Incrementally generate plausible hypotheses, evaluate them, then filter out those that do not make sense. This can be viewed as an epistemic analogue to recent accounts of open-ended decision making \shortcite{morris2021generating,phillips2019we}.\footnote{While this algorithm has $O(T^2)$ runtime due to rejuvenation, in practice runtime is closer to $O(T)$ due to reuse of likelihood computations in L6. A slight variant can guarantee $O(T)$ runtime by forgetting previous observations \shortcite{beronov2021sequential}.}

\begin{figure*}[t]
    \begin{subfigure}[b]{\textwidth}
    \includegraphics[width=\textwidth]{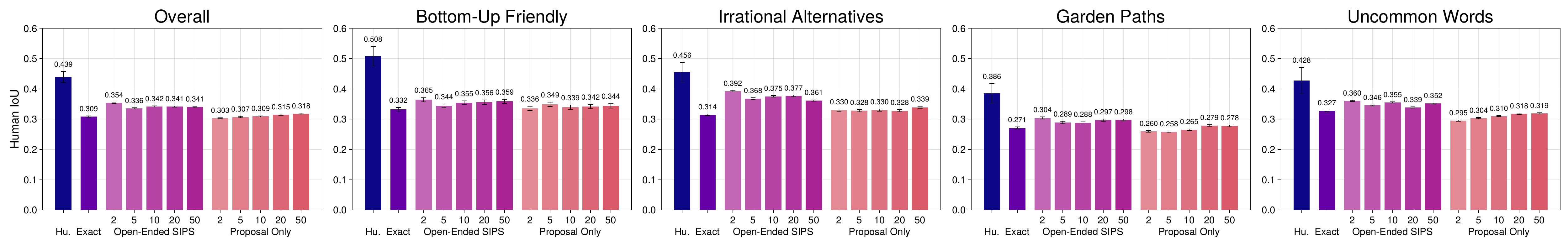}
    \caption{\textbf{Intersection over Union} (IoU) between model inferences and the mean human distribution of guesses.}
    \label{fig:iou}
    \end{subfigure}
    \begin{subfigure}[b]{\textwidth}
    \includegraphics[width=\textwidth]{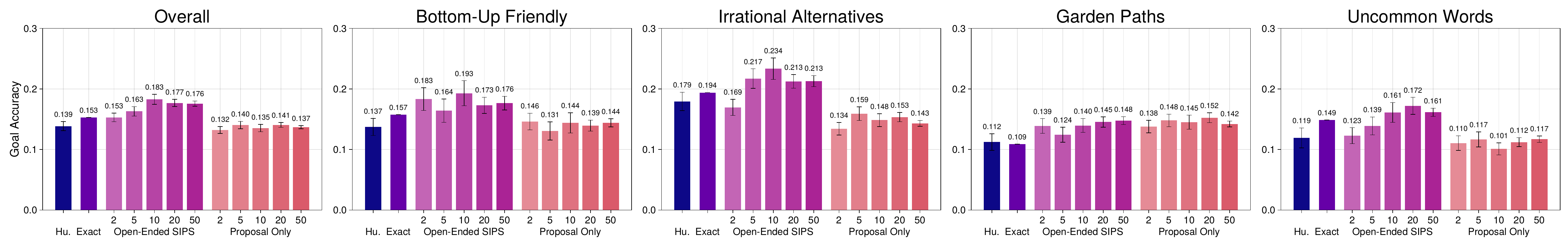}
    \caption{\textbf{Average goal accuracy} (posterior probability of true goal) across humans, models, and conditions.}
    \label{fig:accuracy}
    \end{subfigure}
    \caption{Human similarity and accuracy of goal inference models, measured in terms of \textbf{(a)} IoU with mean human inferences,
    and \textbf{(b)} average goal accuracy. Each bar corresponds to a model (and sample size $N$), while each column is a condition. We computed human-human IoU through repeated sampling of 50-50 splits. Error bars denote 95\% confidence intervals.}
    \label{fig:results}
\end{figure*}

\begin{figure*}[p]  
    \centering
    \begin{subfigure}[b]{0.48\textwidth}
    \includegraphics[width=\textwidth]{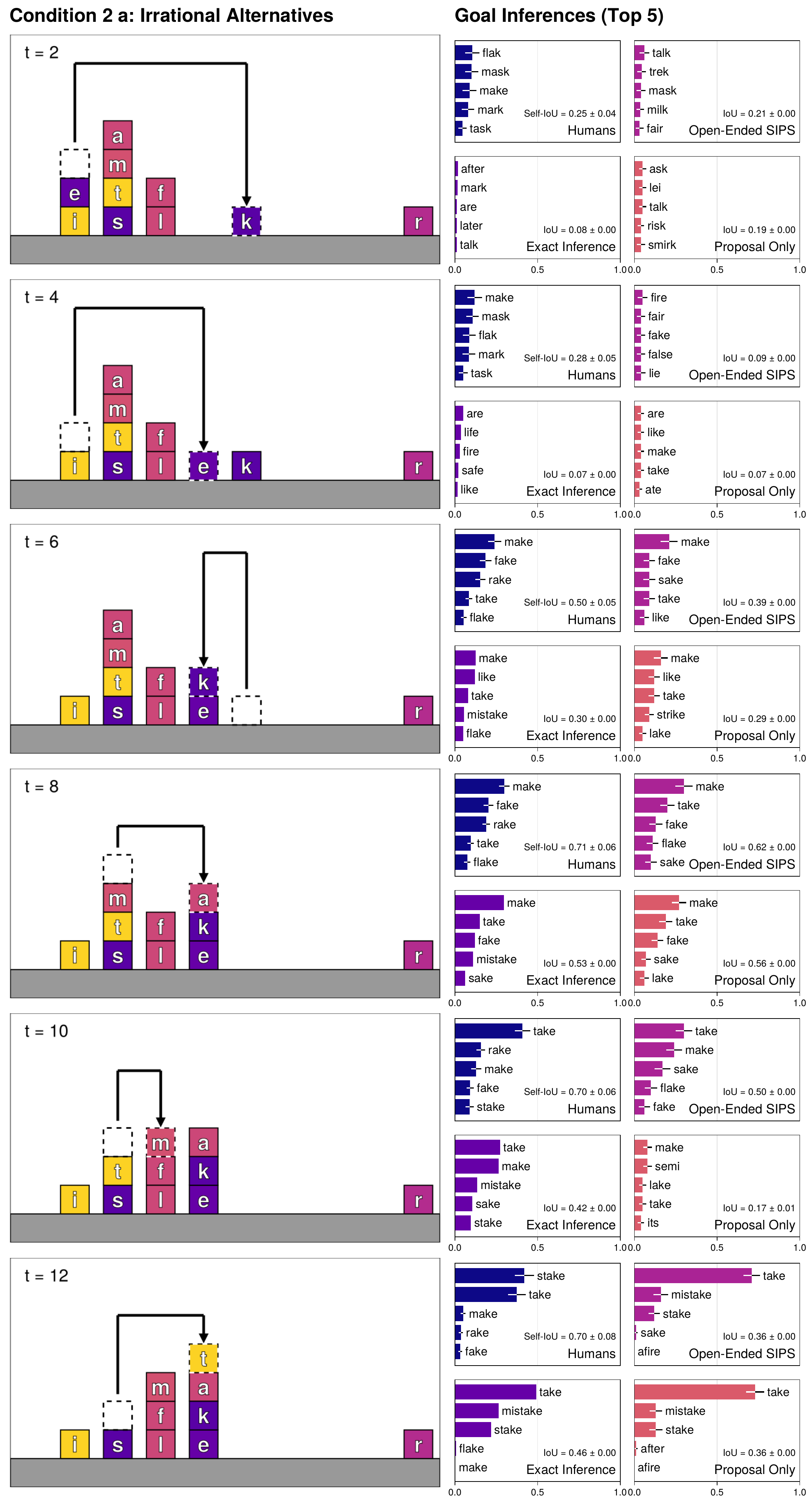}
    \caption{Irrational Alternatives scenario: \textbf{\textsf{stake}}}
    \end{subfigure}
    \begin{subfigure}[b]{0.48\textwidth}
    \includegraphics[width=\textwidth]{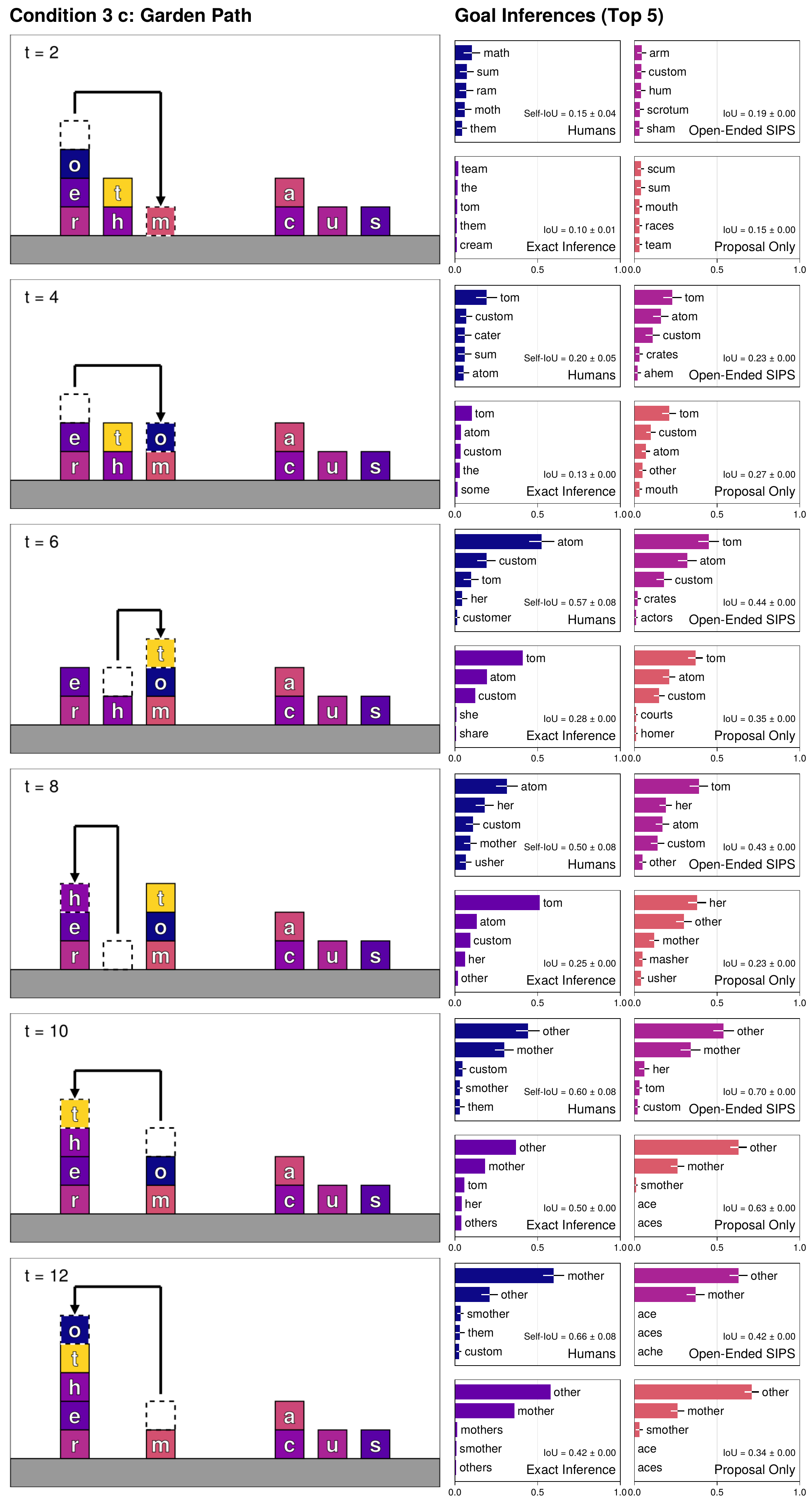}
    \caption{Garden Path scenario: \textbf{\textsf{mother}}}
    \end{subfigure}
    \caption{Step-by-step inference results on two illustrative Block Words scenarios. On the left, we show the sequence of actions, and on the right, the 5 most probable goals at each step for humans and our models ($N=2$ for the sampling-based methods), averaged across humans and algorithm runs (error bars reflect the standard error). In \textbf{(a)}, only the bottom-up proposal fails to infer that \textbf{\textsf{m}} being stacked on \textbf{\textsf{f}} at $t=10$ implies that \textbf{\textsf{make}} and \textbf{\textsf{fake}} are less likely than \textbf{\textsf{take}}. In \textbf{(b)}, both open-ended SIPS and humans exhibit sticky inferences at $t=8$, assigning high weight to \textbf{\textsf{atom}} and \textbf{\textsf{custom}} as guesses as a result of the garden path trajectory. In contrast, the bottom-up proposal displays a recency bias since it does not store previous guesses.}
    \label{fig:storyboards}
\end{figure*}

\section{Experiments}

We evaluated open-ended SIPS as a model of human goal inference on a set of 16 scenarios in Block Words, a variant of the classic Blocksworld domain where the goal is to \emph{infer} the word that an agent is spelling by stacking a tower of lettered blocks. In contrast to previous Block Words tasks  \shortcite{ramirez2009plan,ramirez2010probabilistic,alanqary2021modeling,chandra2023inferring}, we did not specify a fixed set of 5 to 20 goal words. Instead, we told participants that the goal might be any English word between 3 to 8 letters long, with the implied restriction that the word had to be spelled out of the available blocks.

\vspace{-6pt}
\paragraph{Structure \& Design}
Participants were first shown the initial layout of the blocks. They could then advance the scenario, watching several actions play out as an animated video. The video would then pause at a judgment point, giving participants time to guess the word being spelled via text box entry. Participants could \emph{add} as many guesses as they liked, and also \emph{remove} any previous guesses that they no longer considered likely. They could then advance to the next judgment point, continuing in this way until the end of the scenario. Each participant was presented 8 out of the 16 scenarios, after first completing a tutorial and a comprehension quiz. To incentivize high quality responses, we paid participants a reward based on the accuracy of their guesses ($\$0.1/n$ for every correct answer out of $n$ guesses), and presented the bonus point breakdown after they completed each scenario.

\paragraph{Experimental Conditions}

To tease apart the predictions of our model from those that would be made by either exact Bayesian inference or pure bottom-up proposals, we designed our 16 scenarios to fall into one of four conditions:

\emph{\textbf{Bottom-Up Friendly.}} Words are stacked more-or-less linearly, such that it is sufficient to guess words that complete either the most recently stacked tower, or any partial word.

\emph{\textbf{Irrational Alternatives.}} Blocks are stacked so that some bottom-up guesses are made irrational, like our \block{\textsf{p}}\block{\textsf{i}}\block{\textsf{n}}\block{\textsf{k}} example from Figure \ref{fig:example}.

\emph{\textbf{Garden Paths.}} Cases where bottom-up guessing suggests a plausible but misleading interpretation of the first few actions, which turn out to be merely instrumental for the true goal.

\emph{\textbf{Uncommon Words.}} The true goal is either a longer or more uncommon word (aft, chump, wizard, banish), which people and bottom-up proposals might find difficult to initially guess. Otherwise similar to the first condition.

\vspace{-6pt}
\paragraph{Participants}

We recruited 100 US participants fluent in English via Prolific (mean age 39.4; 44 women, 54 men, 2 non-binary), such that every scenario was completed by 50 individuals. Participants were paid US\$15/hr along with the bonus described earlier. Familiarity with word games varied, with 17 reporting that they played word games daily, 22 weekly, 30 every 1-2 months, 11 yearly, and 20 almost never.

Despite comprehension checks, a subset of participants did not follow instructions correctly, either because they never updated their guesses (36 out of 800 scenario responses), or only added guesses without removing them (139 out of 800). As such, we excluded such responses from our analysis. 

\begin{figure*}[t]
\centering
\hspace{1.5cm}\includegraphics[width=0.8\textwidth]{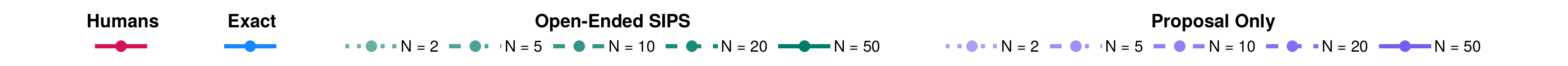}

\begin{subfigure}[t]{0.325\textwidth}
\includegraphics[width=\textwidth]{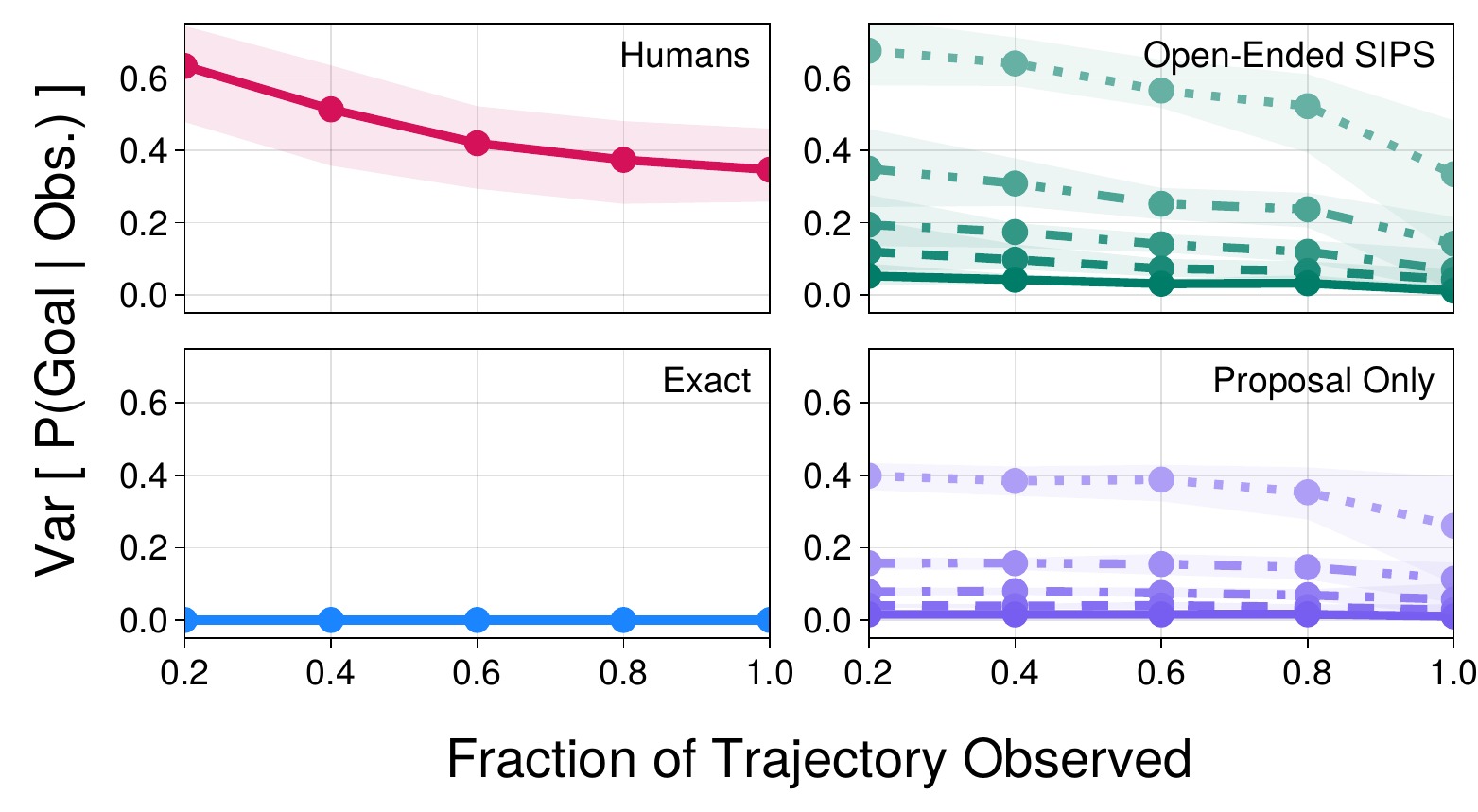}
\caption{\textbf{Total variance} of the goal probability estimates produced by each method vs. humans, averaged over scenarios.}
\label{fig:variance}
\end{subfigure}
\hfill
\begin{subfigure}[t]{0.325\textwidth}
\includegraphics[width=\textwidth]{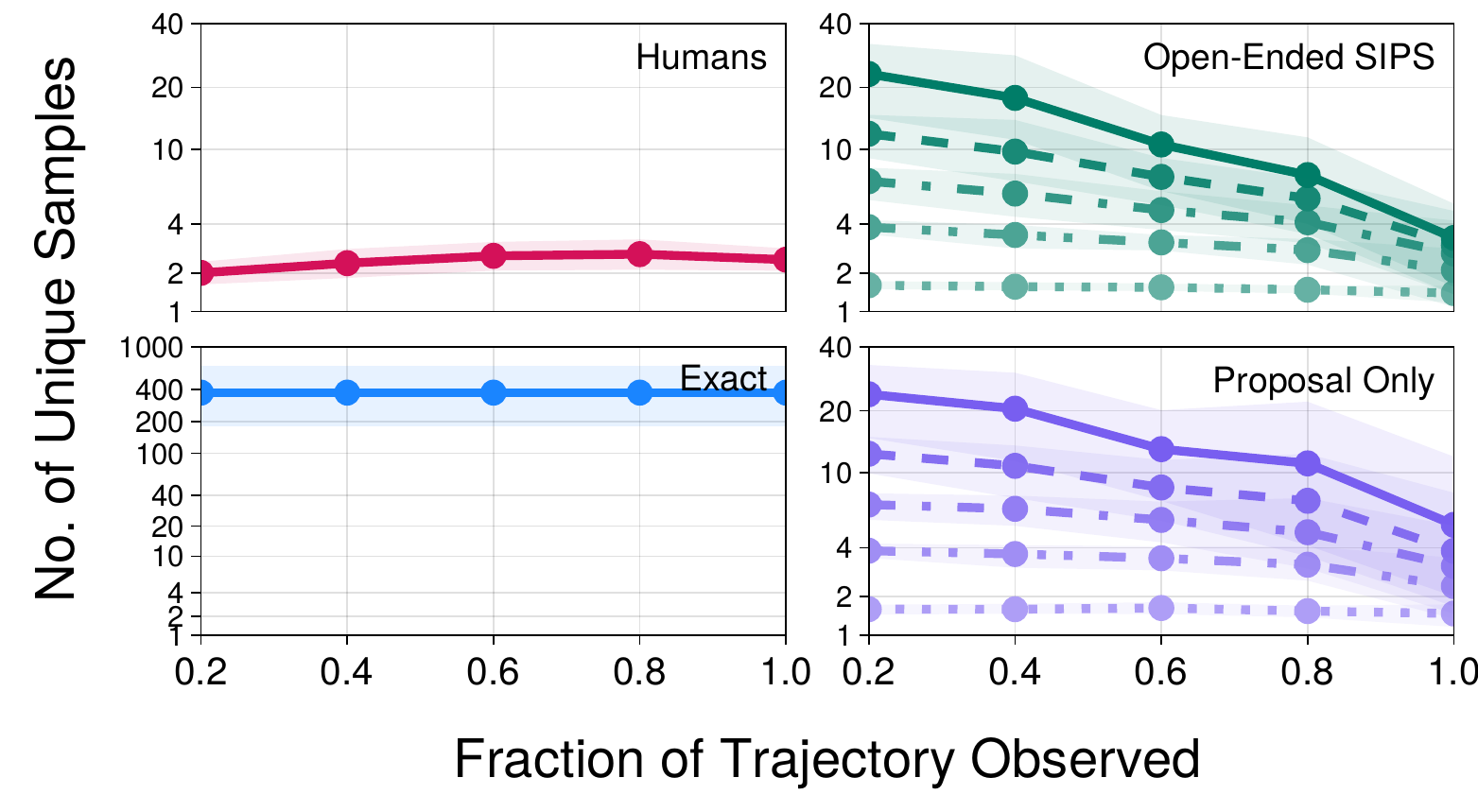}
\caption{\textbf{Sample efficiency} of inference methods over time, measured by the number of \emph{unique} tracked hypotheses (averaged over scenarios).}
\label{fig:sample-efficiency}
\end{subfigure}
\hfill
\begin{subfigure}[t]{0.325\textwidth}
\includegraphics[width=\textwidth]{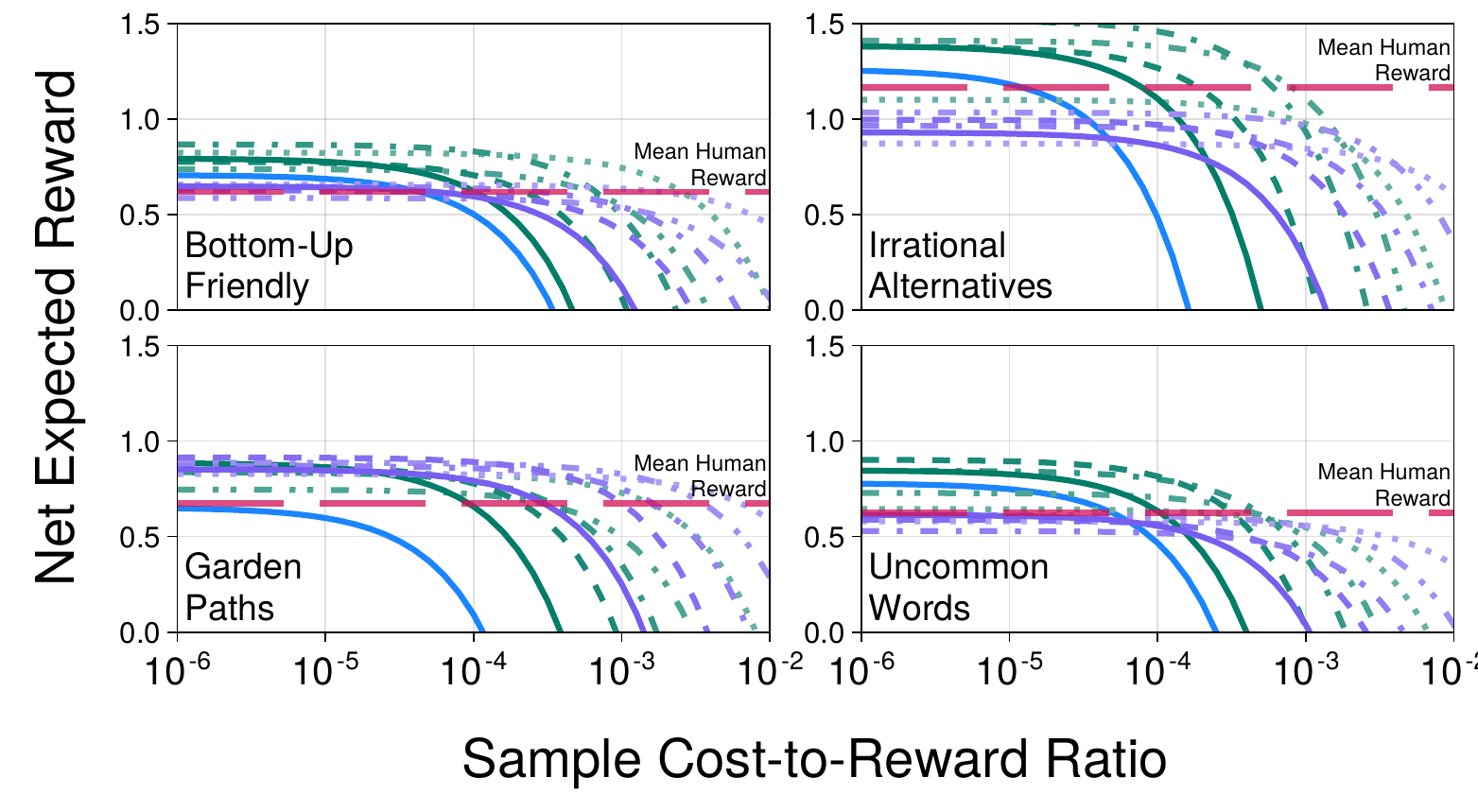}
\caption{\textbf{Expected reward minus cognitive cost}, as a function of the ratio of sample cost to the reward of accurate goal inference.}
\label{fig:resource-rationality}
\end{subfigure}
\caption{Response variance, sample efficiency, and cognitive cost trade-offs vs. humans. Ribbons show 10th-90th quantiles.}
\label{fig:algprops}
\end{figure*}

\vspace{-6pt}
\paragraph{Model Configuration}

We implemented open-ended SIPS using the particle filtering extension of the Gen.jl probabilistic programming framework \shortcite{zhi2020gen,cusumano2019gen}, and the Blocksworld domain in the Planning Domain Definition Language \cite{zhi2022pddl,mcdermott1998pddl}. We fit parameters via grid search to improve model similarity with humans as measured by the intersection over union (IoU) between distributions (see Appendix), which gave an inverse temperature of $\beta = 1.0$, planning budget of $B = 100$, and prior $P(g)$ fitted to tempered word frequencies from the \texttt{wordfreq} library \cite{speer2016wordfreq}, using the \texttt{3of6game} word list as our dictionary \cite{beale201612dicts}. We ran open-ended SIPS with $N \in \{2, 5, 10, 20, 50\}$ particles, taking the mean and variance over $M = \max(10, 200/N)$ trials. We describe the proposals $Q(g|s_t,a_t)$ in the next section.
\vspace{-9pt}

\paragraph{Alternative Models}

As alternatives to our hybrid SMC model, we tested both (i) exact inference via fully enumerative Bayesian inverse planning over all valid English goal words (145--807 words, depending on the scenario) and (ii) pure bottom-up sampling using a subgoal-conditioned proposal. Exact inference was implemented using the same model parameters as open-ended SIPS, except that all goals were considered as hypotheses from the outset.

For the bottom-up proposal, we sampled complete words $g$ by conditioning an $n$-gram model on some partial word that can be stacked from blocks in the current state $s_t$. We used $n=5$, fitting the $n$-gram on the same tempered word frequencies used for the prior $P(g)$. To decide which partial word to complete, the proposal first considers the tower stacked by the last action $a_t$, sampling a completion if it is sufficiently word-like (as determined by the $n$-gram model). If not, it considers if the last block was moved because the agent intended to reach some block \emph{underneath} it, and tries to form a word with one of those blocks. If no way of using those blocks is sufficiently word-like, the proposal samples a random tower in state $s_t$ (weighted by how word-like it is), then samples a completion. Other proposals are discussed in the Appendix.

\subsection{Results}

We analyzed human responses and model outputs by comparing them in terms of distribution similarity (Fig. \ref{fig:iou}), average accuracy (Fig. \ref{fig:accuracy}), step-by-step inferences (Fig. \ref{fig:storyboards}), response variance (Fig. \ref{fig:variance}), sample efficiency (Fig. \ref{fig:sample-efficiency}), and resource rationality (Fig. \ref{fig:resource-rationality}). Additional results (e.g. accuracy vs. runtime) and sensitivity analyses are in the Appendix.

\textbf{Open-ended SIPS is most similar to human inferences across all conditions.} As we predicted, human inferences showed the highest similarity with open-ended SIPS (IoU = 0.33--0.36 for all $N$) compared to exact inference (IoU = 0.31) or bottom-up guessing (IoU = 0.30--0.32), with $N = 2$ samples being the most similar. Notably, open-ended SIPS was more similar to humans in the \emph{Irrational Alternatives} condition, with both achieving considerably higher accuracy than the bottom-up only heuristic, indicating that humans indeed engage in inverse planning. Our model was also more similar to humans than exact inference, especially in the \emph{Bottom-Up Friendly} and \emph{Garden Path} conditions, consistent with the hypothesis that humans engage in bottom-up sampling.

\textbf{Step-by-step human inferences are best matched by open-ended SIPS.} The step-by-step comparisons in Figure \ref{fig:storyboards} help to elucidate these aggregate findings. On one hand, open-ended SIPS and the proposal-only model make initial guesses that are biased towards words that complete the first few stacked letters, whereas the exact posterior is much more uncertain. On the other hand, humans account for the rationality of the observed actions when drawing inferences (e.g. Figure \ref{fig:storyboards}(a), $t=10$), just like our exact and approximate Bayesian inverse planning algorithms.

\textbf{Our model's algorithmic properties best explain human variance and guess counts.} In Figure \ref{fig:algprops}, we compare the \emph{algorithmic} properties of the models. Human variance was best matched by open-ended SIPS with $N = 2$. Bottom-up proposals had lower variance, and did not prune samples as effectively as sample-matched counterparts. Exact inference is zero-variance, but at the cost of tracking drastically more hypotheses. As such, it was dominated by open-ended SIPS in terms of net reward when accounting for cognitive costs (Fig. \ref{fig:resource-rationality}). The comparison with pure bottom-up proposals was more nuanced. If reweighting a sample via inverse planning is costly enough, pure bottom-up guessing can be more resource-rational \cite{lieder2020resource}. However, there is a large range of cost-ratios where it pays to do inverse planning. Since humans attained more reward than all proposal-only baselines in the \emph{Irrational Alternatives} condition, this suggests that they indeed find inverse planning worthwhile.

\section{Discussion}

In comparison to alternative models, our sampling-based account of open-ended goal inference is best supported on both empirical and theoretical grounds, providing an algorithmically plausible explanation for the speed and flexibility of human goal inference.  Still, our experiments find that humans remain more similar to themselves (IoU = 0.44) than our best-fitting model (IoU = 0.35). Part of this might be explained by the discrepancy between the statistics of how humans guess in word games versus the text corpus frequencies that inform our model. This could be addressed by deriving a prior and proposal from human guesses. Humans also appear to exhibit \emph{stickier} inferences in garden path cases, whereas open-ended SIPS tends to avoid them when run with larger values of $N$ by proposing new goals at every step. This suggests that humans may be \emph{adaptive} in deciding when to rejuvenate their hypotheses \shortcite{del2012adaptive,elvira2016adapting}. Finally, unlike our model, humans might forget older observations, becoming more inaccurate, but also more efficient at inference. SMC algorithms that selectively forget past observations could mimic this \shortcite{beronov2021sequential}.

Another open question is how bottom-up sampling can be made more general. In future work, we plan to explore how the statistics of co-occurring subgoals can be distilled from web-scale language models \shortcite{west2022symbolic} into domain-specific models for rapid hypothesis generation. These statistics might be augmented by static analysis of environment models, automatically determining which subgoals are instrumental for other goals \shortcite{blum1997fast}. Such mechanisms for flexible domain adaptation could provide an even richer picture of how we contend with the infinitude of ends that others pursue, even in the face of our very finite means.

\section{Acknowledgements}
This work was funded in part by the DARPA Machine Common
Sense, AFOSR, and ONR Science of AI programs, along with the
MIT-IBM Watson AI Lab and gifts from the Aphorism Foundation and the
Siegel Family Foundation. Tan Zhi-Xuan is supported by an Open
Philanthropy AI Fellowship.

\section{Code and Data Availability}

Code and data for this paper can be found at the accompanying OSF repository: \url{https://osf.io/bygwm/}

\bibliographystyle{apacite}

\setlength{\bibleftmargin}{.125in}
\setlength{\bibindent}{-\bibleftmargin}

\bibliography{paper}

\newpage

\renewcommand\thefigure{A\arabic{figure}}
\setcounter{figure}{0}

\renewcommand\thetable{A\arabic{table}}
\setcounter{table}{0}

\section{Appendix}

\subsection{Experiment Interface}

\begin{figure}[h]
    \centering
    \includegraphics[width=\columnwidth]{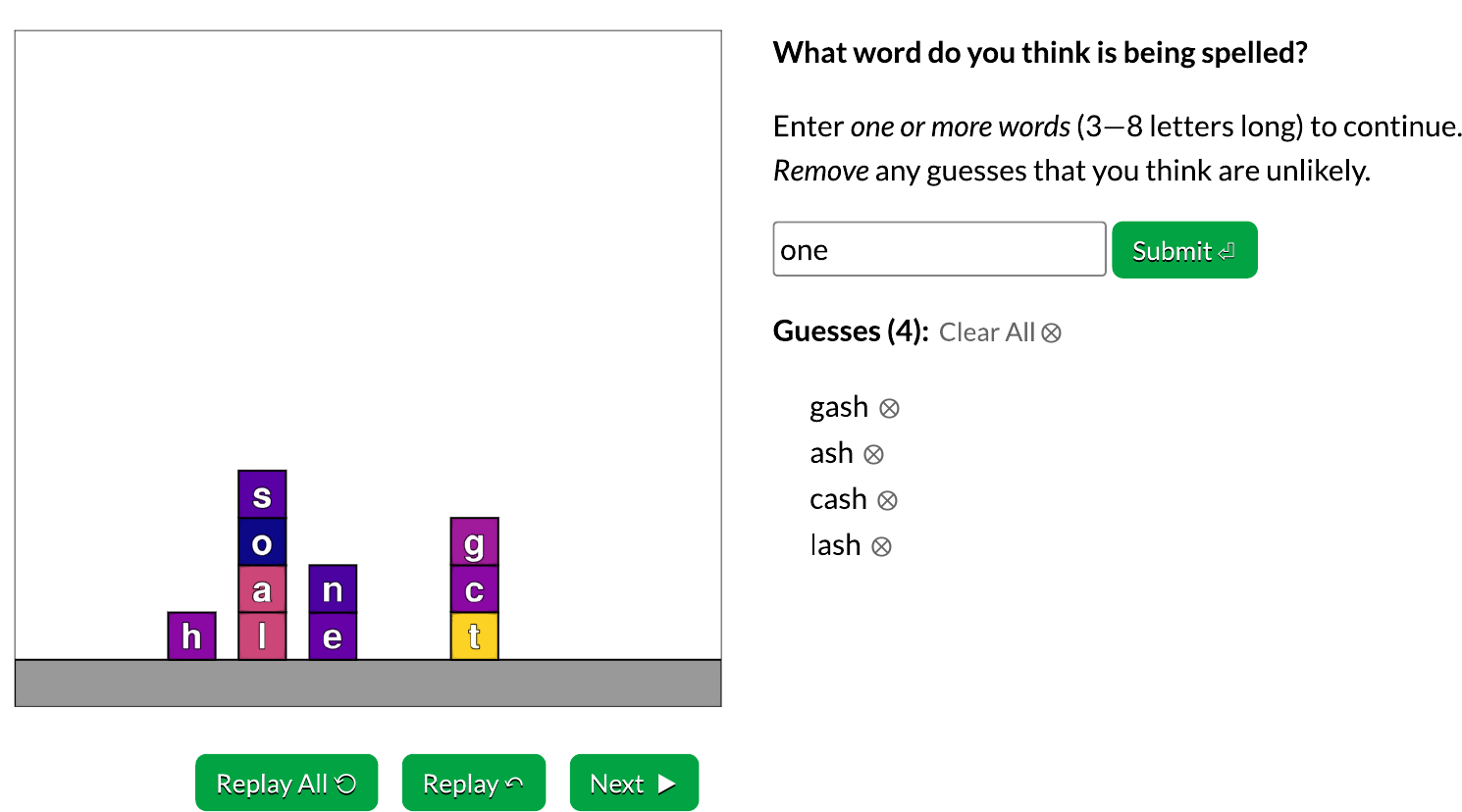}
    \caption{Interface for our open-ended goal inference task.}
    \label{fig:interface}
\end{figure}

The web interface used by participants is shown in Figure \ref{fig:interface}. At each judgment point, participants typed their guesses into the text box, which validated whether the guess was between 3 and 8 characters and used only the letters that were available. Participants could also remove their guesses by clicking the $\bigotimes$ symbol next to each guess. The list of guesses was converted into a probability distribution by assigning a probability of $1/n$ to each word among the $n$ guesses. Participants could rewatch the most recent segment of the animation by pressing the \emph{Replay} button, or rewatch the whole animation up to the judgment point by pressing the \emph{Replay All} button. This interface is accessible at \url{https://block-words.web.app/?local=true}.

\subsection{Model Fitting and Sensitivity Analysis}

Our model of open-ended goal inference is characterized by two sets of parameters: The parameters of the \emph{generative model} $P(g, \pi_{1:t}, s_{0:t}, a_{1:t})$, and the parameters of the \emph{inference algorithm} which approximates $P(g | s_{0:t}, a_{1:t})$. We fit the parameters of the generative model across the following ranges:
\begin{itemize}[itemsep=0pt]
    \item Goal prior word temperature $T_w \in \{1, 2, 4, 8, 16\}$
    \item Inverse temperature $\beta \in \{\frac{1}{4}, \frac{1}{2}, 1, 2, 4\}$
    \item Planning budget $B \in \{5, 10, 20, 50, 100, 200, 500\}$
    \item Replanning cadence $\Delta t \in \{1, 2\}$
    \item RTHS search strategy $\sigma =$ A* or BFS
\end{itemize}
$T_w$ controls tempering of the \texttt{wordfreq}-derived word frequencies used for the goal prior $P(g)$, and $\beta$ controls the optimality of action selection. $B$ is the planning budget for real-time heuristic search (RTHS) algorithm, $\Delta t$ is the number of timesteps between each call to RTHS that updates the policy $\pi_t$, and $\sigma$ controls how nodes are expanded by RTHS, which is done either via A* search around each neighbor of the current state $s_t$ (guided by the FF heuristic as the default $\hat Q_{\pi_t}$ value) as in LSS-LRTA* \shortcite{koenig2009comparing}, or via breadth-first search (BFS) around the current state $s_t$, as in LRTA*-LS \shortcite{hernandez2007improving}.

\begin{figure}[t]
    \centering
    \includegraphics[width=\columnwidth]{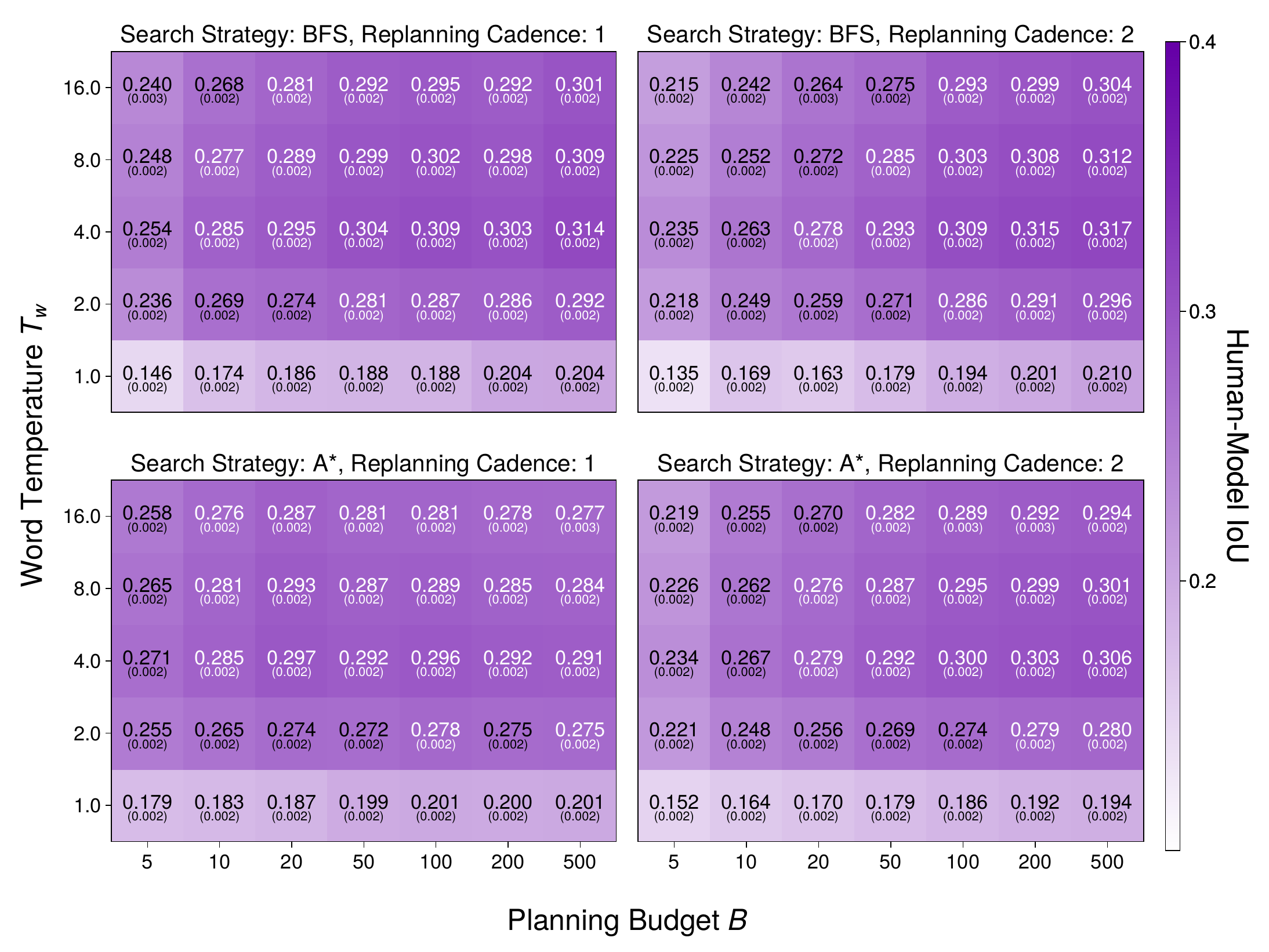}
    \caption{Human-model similarity (IoU) across generative model parameters when using exact Bayesian inference.}
    \label{fig:model-param-heatmap}
    \vspace{-6pt}
\end{figure}

For the inference algorithm, we fit these parameters:
\begin{itemize}[itemsep=0pt]
    \item $n$-gram word temperature $T_w \in \{1, 2, 4, 8, 16\}$
    \item $n$-gram termination bias $\epsilon \in \{0, 0.05, 0.1, 0.15, 0.2, 0.25\}$
    \item Bottom-up proposal strategy $Q \in \{\textsc{last-and-next}, ...\}$
    \item Number of samples $N \in \{2, 5, 10, 20, 50\}$
\end{itemize}
$T_w$ tempers the word frequencies used to fit the $n$-gram model for the bottom-up proposal $Q$, and is matched to be the same value used for the goal prior $P(g)$. To capture the difficulty of guessing longer words, we modified the $n$-gram to have an additional $\epsilon$ probability of terminating after each character. For simplicity, we fixed the context length of the $n$-gram model to $n = 5$. Various ways of implementing the bottom-up proposal $Q$ are discussed in the next section. We also vary the number of particles $N$ used by open-ended SIPS.

\textbf{Fitting procedure.} Model fitting proceeded in two stages. We first fit the generative model parameters to improve similarity with humans, using \emph{exact} inference to factor out stochasticity or performance issues in the inference algorithm from the quality of the generative model itself. Instead of Pearson's correlation coefficient (commonly used in other BToM studies), we used the intersection-over-union between human and model distributions (i.e. the Jaccard index) as our similarity metric, since it does not consider two probability vectors similar just because they both contain many zeros. Having determined values of $B = 100$, $\Delta t = 2$ and $\sigma = \text{BFS}$ that led to the most similarity with humans under the constraint of a reasonable runtime, we then fit the parameters of the open-ended SIPS algorithm. The best fitting inference parameters were $T_w = 4$ (which was matched with the goal prior's $T_w$), $\epsilon = 0.05$, $Q = \textsc{last-and-next}$, and $N = 2$.

\textbf{Generative model sensitivity analysis.} Figure \ref{fig:model-param-heatmap} shows how similarity with humans varies across generative model parameters when using exact Bayesian inference. A higher planning budget $B$ leads to a stronger fit, showing the importance of computing a good estimate of the agent's policy via planning. Interestingly however, the more informed search strategy, A*, led to slightly worse fits, suggesting that humans may not be explicitly modeling other's detailed search processes when performing inverse planning over large numbers of goals. Human similarity also improved when using a tempered word prior with $T_w = 4.0$, whereas using the raw corpus frequencies as the prior ($T_w = 1.0$) led to substantially poorer fits. In other words, people's intuitions for what words are likely in a word game are substantially broader than everyday word usage statistics.

\begin{figure}[t]
    \centering
    \includegraphics[width=\columnwidth]{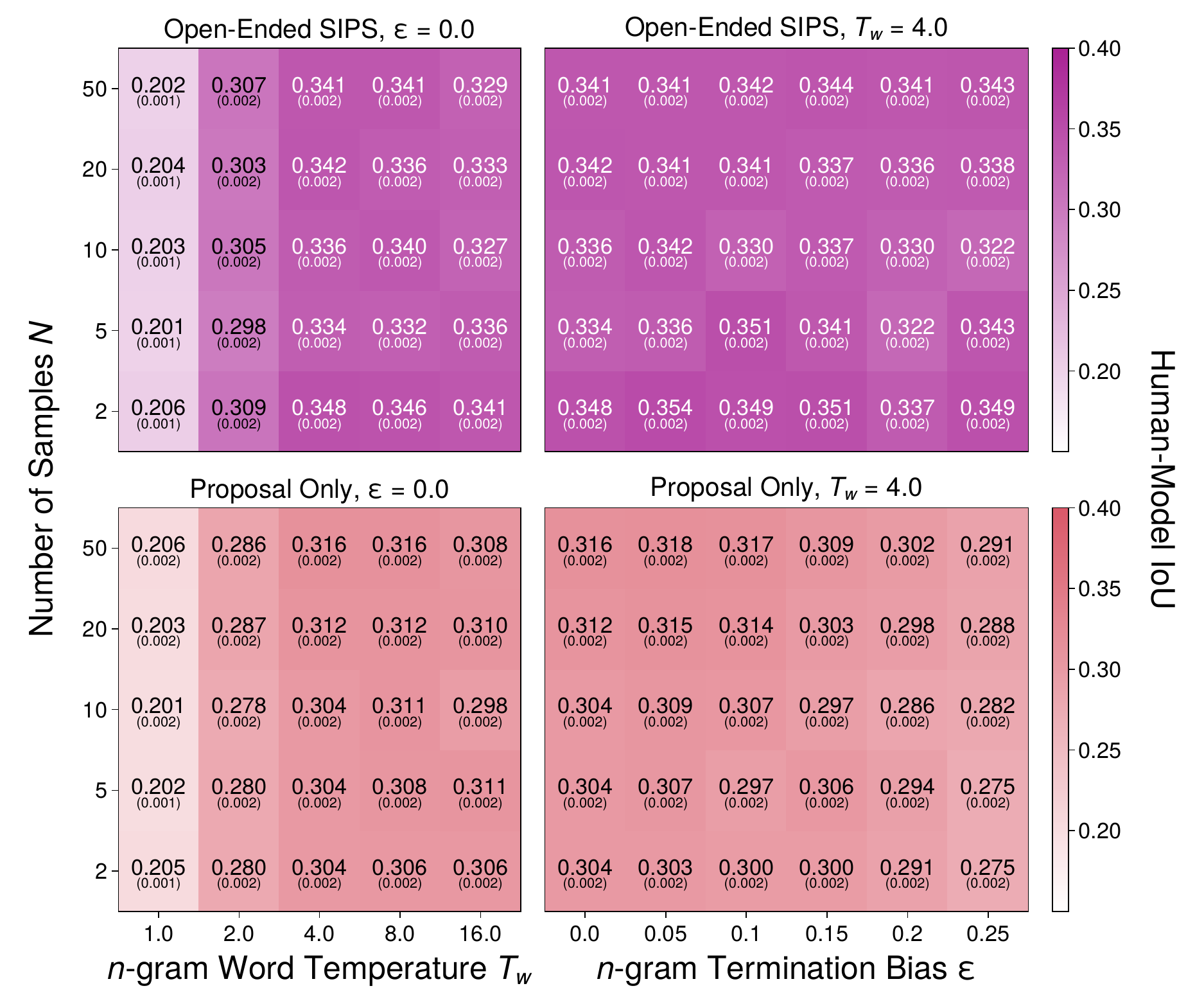}
    \caption{Human-model similarity (IoU) across inference parameters for open-ended SIPS and the bottom-up proposal.}
    \label{fig:inference-param-heatmap}
    \vspace{-9pt}
\end{figure}

\textbf{Inference algorithm sensitivity analysis.} Figure \ref{fig:inference-param-heatmap} shows how human-model IoU varies with different inference parameters for both open-ended SIPS and the proposal-only baseline. Again, untempered word frequencies ($T_w = 1$) lead to a poor fit, which cannot be overcome even with a highe sample count $N$. Intermediate tempering ($T_w = 4$) leads to the best fit, capturing the broader distribution of human word guesses. The termination bias $\epsilon$ has a less pronounced effect, with the best value ($\epsilon = 0.05)$ capturing the additional difficulty of proposing long words. Open-ended SIPS dominates the bottom-up proposal for almost all parameter settings.

\subsection{Bottom-Up Proposals}

\begin{table}[t]
\scriptsize
\centering

\begin{subtable}[b]{0.9\columnwidth}
\begin{tabular}{@{}llllll@{}}
\toprule
\textbf{Open-Ended SIPS}                         & \multicolumn{5}{c}{\textbf{Human-Model Similarity (IoU)}} \\
\textbf{Proposal Strategy} $Q$                   & $N=2$   & $N=5$   & $N=10$   & $N=20$  & $N=50$  \\ \midrule
\textsc{any-tower}     & 0.294    & 0.310   & 0.297   & 0.314   & 0.319   \\
\textsc{last-tower}    & 0.348    & 0.340   & 0.317   & 0.339   & 0.344   \\
\textsc{next-tower}    & 0.302    & 0.304   & 0.306   & 0.309   & 0.329   \\
\textsc{last-and-next} & \textbf{0.354}    & 0.336   & 0.342   & 0.341   & 0.341   \\ \midrule
                         & \multicolumn{5}{c}{\textbf{Goal Accuracy}} \\ \midrule
$\textsc{any-tower}$     & 0.115  & 0.162  & 0.156  & 0.159  & 0.160   \\
$\textsc{last-tower}$    & 0.163  & 0.171  & \textbf{0.184}  & 0.174  & 0.176  \\
$\textsc{next-tower}$    & 0.118  & 0.157  & 0.160   & 0.170   & 0.165  \\
$\textsc{last-and-next}$ & 0.153  & 0.163  & \textbf{0.183}  & 0.177  & 0.176  \\ \bottomrule
\end{tabular}
\vspace{-3pt}
\subcaption{Open-Ended SIPS}
\end{subtable}
\vspace{3pt}

\begin{subtable}[b]{0.9\columnwidth}
\begin{tabular}{@{}llllll@{}}
\toprule
\textbf{Bottom-Up}                         & \multicolumn{5}{c}{\textbf{Human-Model Similarity (IoU)}} \\
\textbf{Proposal Strategy} $Q$                   & $N=2$   & $N=5$   & $N=10$   & $N=20$  & $N=50$  \\ \midrule
\textsc{any-tower}     & 0.211    & 0.218   & 0.217   & 0.224   & 0.228   \\
\textsc{last-tower}    & 0.300    & 0.301   & 0.301   & 0.306   & 0.310   \\
\textsc{next-tower}    & 0.226    & 0.223   & 0.221   & 0.231   & 0.237   \\
\textsc{last-and-next} & 0.303    & 0.307   & 0.309   & 0.315   & \textbf{0.318}   \\ \midrule
                         & \multicolumn{5}{c}{\textbf{Goal Accuracy}}     \\ \midrule
$\textsc{any-tower}$     & 0.067 & 0.071 & 0.072 & 0.069 & 0.070  \\
$\textsc{last-tower}$    & \textbf{0.141} & 0.134 & 0.139 & 0.138 & 0.134 \\
$\textsc{next-tower}$    & 0.073 & 0.074 & 0.068 & 0.072 & 0.074 \\
$\textsc{last-and-next}$ & 0.132 & 0.140  & 0.135 & \textbf{0.141} & 0.137 \\ \bottomrule
\end{tabular}
\vspace{-3pt}
\subcaption{Proposal-Only Baseline}
\end{subtable}
\vspace{-6pt}

\caption{Effect of bottom-up proposal strategy $Q$ on human-model similarity (IoU) and goal accuracy.}
\label{tab:proposal-comparison}
\vspace{-9pt}
\end{table}

For bottom-up sampling, we experimented with proposals $Q(g|s_t,a_t)$ of varying degrees of sophistication. In all of these proposals, we sample complete words $g$ by conditioning an $n$-gram language model on some partial word that can be stacked from the blocks in the current state $s_t$. However, this still leaves undetermined \emph{which} partial words, or subgoals, to consider. We implemented the following strategies for selecting partial words to complete:

\vspace{3pt}
\textbf{\textsc{any-tower}} samples a random block tower $\tau$ in state $s_t$, then tries to complete it into a full word. The probability of sampling a tower is proportional to how word-like the tower is --- i.e., how probable it is according the $n$-gram model.

\textbf{\textsc{last-tower}} tries to sample a word that completes the tower \emph{most recently} stacked by action $a_t$. However, if this tower $\tau$ is not sufficiently word-like (or if $a_t$ is not a stacking action), the proposal falls back to \textsc{any-tower} instead. This is implemented by comparing the probability $p_\text{last}$ of the most recently stacked tower under the $n$-gram against the probability $p_\text{rand}$ of an equally tall tower of random blocks, then deciding to complete the last tower with probability $\frac{p_\text{last}}{p_\text{last} + p_\text{rand}}$.

\vspace{3pt}
\textbf{\textsc{next-tower}} focuses on cases where the agent is unstacking a tower of blocks in order to reach a block in that tower. The proposal considers all ways of using a block in the most recently unstacked tower to complete some other block tower. One of these candidate towers $\tau$ is selected with probability proportional to how word-like it is, and then a completion is sampled from to the $n$-gram model. If no candidate is word-like enough compared to the probability of selecting a random block, then the proposal defaults to \textsc{any-tower}.

\vspace{3pt}
\textbf{\textsc{last-and-next}} is the most sophisticated of our proposals, which we describe and use in the main text. It is equivalent to \textsc{last-tower}, except it defaults to \textsc{next-tower} if the last action is not a stacking action, or if the last stacked tower is not sufficiently word-like.

\vspace{3pt}
Table \ref{tab:proposal-comparison} shows how these different proposal strategies compare in terms of both human similarity and goal inference accuracy. As expected, the \textsc{last-and-next} proposal best matches human inferences whether it is incorporated into open-ended SIPS (Table \ref{tab:proposal-comparison}(a)) or used on its own (Table \ref{tab:proposal-comparison}(b)), while achieving (close to) the highest accuracy. The \textsc{last-tower} also performs well in this regard, albeit with slightly lower human similarity. In contrast, both \textsc{any-tower} and \textsc{last-tower} fare poorly. Open-ended SIPS is able to make up for their weakness to some degree, showing the value of inverse planning even with a weak proposal.

\begin{figure*}[t]
    \begin{subfigure}[b]{\textwidth}
    \includegraphics[width=\textwidth]{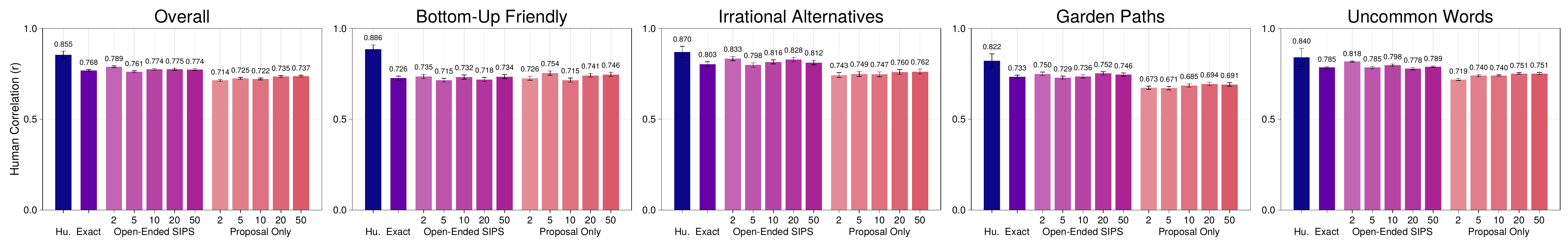}
    \caption{\textbf{Pearson's correlation} $r$ between human inferences and model inferences across conditions.}
    \label{fig:correlation}
    \end{subfigure}
    \begin{subfigure}[b]{\textwidth}
    \includegraphics[width=\textwidth]{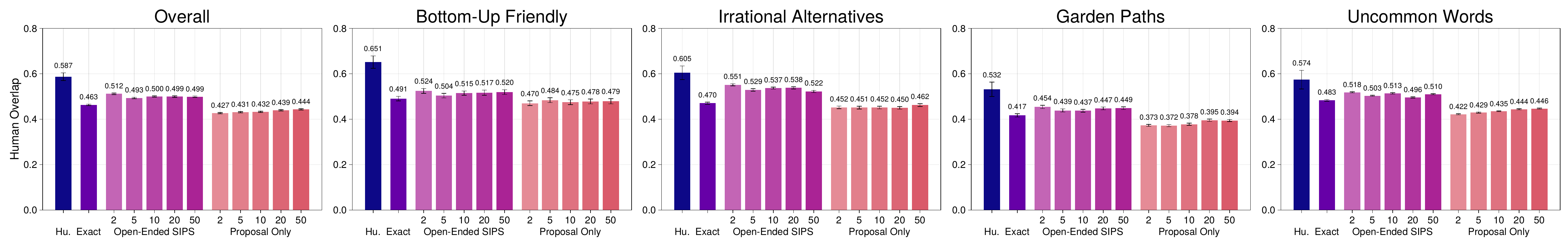}
    \caption{\textbf{Distribution overlap} between human inferences and model inferences across conditions.}
    \label{fig:overlap}
    \end{subfigure}
    \begin{subfigure}[b]{\textwidth}
    \includegraphics[width=\textwidth]{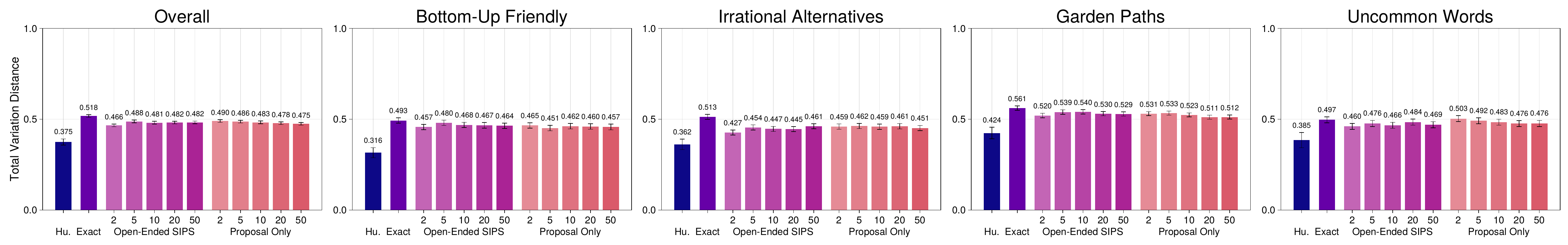}
    \caption{\textbf{Total variation distance} between human inferences and model inferences across conditions (lower is better).}
    \label{fig:total-variation}
    \end{subfigure}
    \caption{Similarity of average human and model goal inferences measured in terms of \textbf{(a)} Pearson's correlation coefficient, \textbf{(b)} distribution overlap ($\sum_{g} \min[P(g), Q(g)]$ for distributions $P$ and $Q$), and \textbf{(c)} total variation distance ($\frac{1}{2} \sum_{g} |P(g), Q(g)|$). As in Figure \ref{fig:results}, error bars denote 95\% CIs, calculated from 1000 bootstrap samples of the distribution of human responses.} 
    \label{fig:similarity-metrics}
\end{figure*}

\textbf{Handling auxiliary randomness.} Note that all of our bottom-up proposals make use of \emph{auxiliary randomness} \shortcite{lew2022recursive} when sampling a tower $\tau$ to complete into a full word $g$. This means that even though we can \emph{sample} from $g \sim Q(g | s_t, a_t)$, we cannot exactly \emph{evaluate} the probability $Q(g | s_t, a_t)$ used in the importance weight. Instead, we can only evaluate $Q(g | s_t, a_t, \tau)$ using the $n$-gram model, which is conditional on the choice of tower $\tau \sim Q(\tau)$. In the context of an SMC algorithm like open-ended SIPS, however, we can use an \emph{unbiased density sampler} of $Q(g | s_t, a_t)$ \shortcite{lew2023probabilistic,lew2022recursive}, which returns both $g$ and a weight $w$ such that $\mathbb{E}_Q[w^{-1} | g] = Q(g | s_t, a_t)^{-1}$. This weight $w$ can then be used as the denominator when computing importance weights in L6 of Algorithm \ref{alg:open-ended-sips}. Using $w = Q(g | s_t, a_t, \tau)$ satisfies this property.

\subsection{Additional Similarity Metrics}

In Figure \ref{fig:similarity-metrics}, we report additional measures of similarity between human goal inferences and model outputs: (a) Pearson's correlation $r$, (b) the overlap coefficient between distributions (a generalized version of recall), and (c) total variation distance (which captures the \emph{maximum} difference in the probability of any event under two distributions). Regardless of the metric, open-ended SIPS is more similar to humans than exact inference or the bottom-up proposal. This is most pronounced in the \emph{Irrational Alternatives} condition.

\subsection{Additional Step-by-Step Comparisons}

\begin{figure*}[p]  
    \centering
    \begin{subfigure}[b]{0.48\textwidth}
    \includegraphics[width=\textwidth]{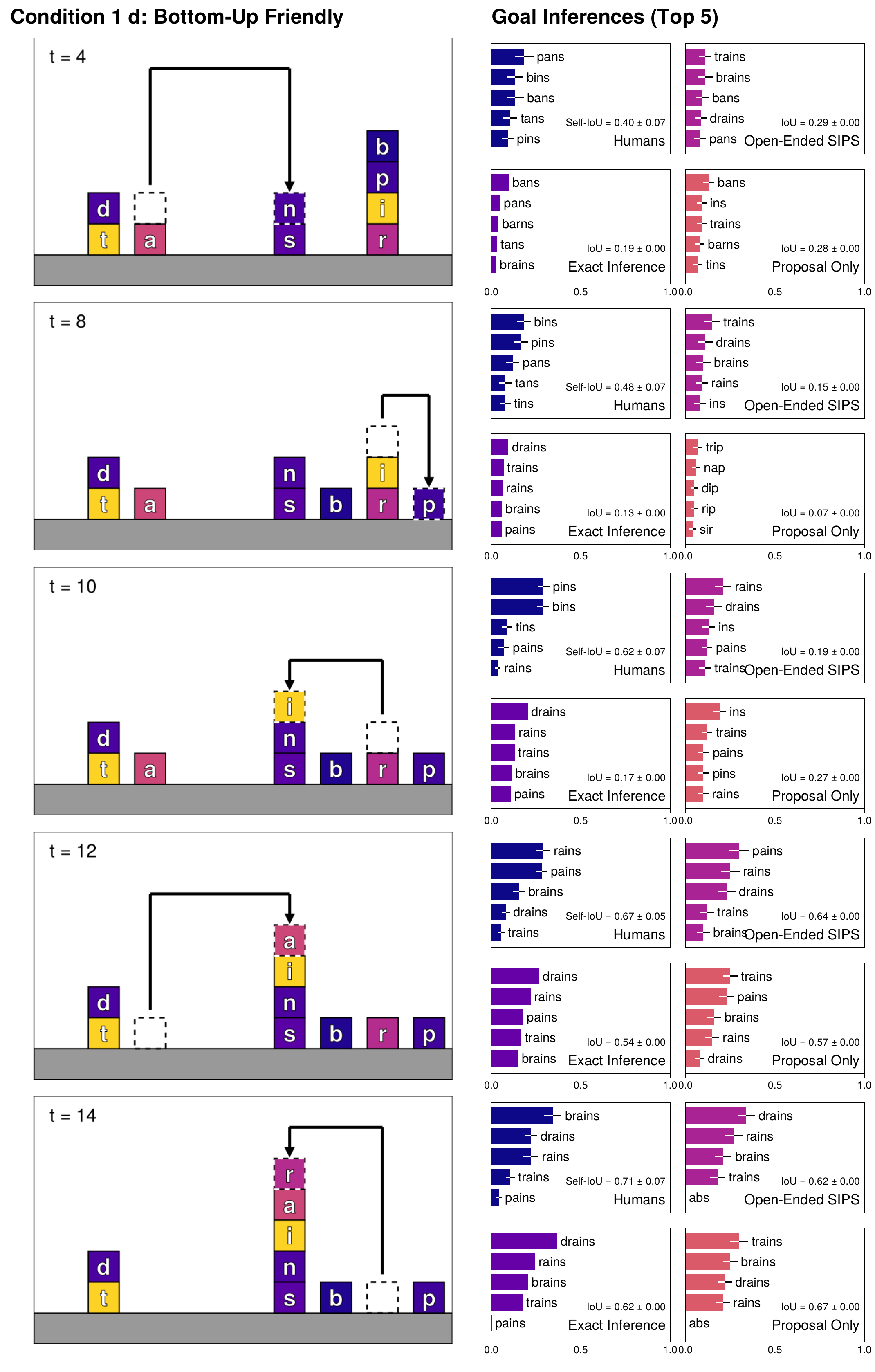}
    \caption{Bottom-Up Friendly scenario: \textbf{\textsf{drains}}}
    \end{subfigure}
    \begin{subfigure}[b]{0.48\textwidth}
    \includegraphics[width=\textwidth]{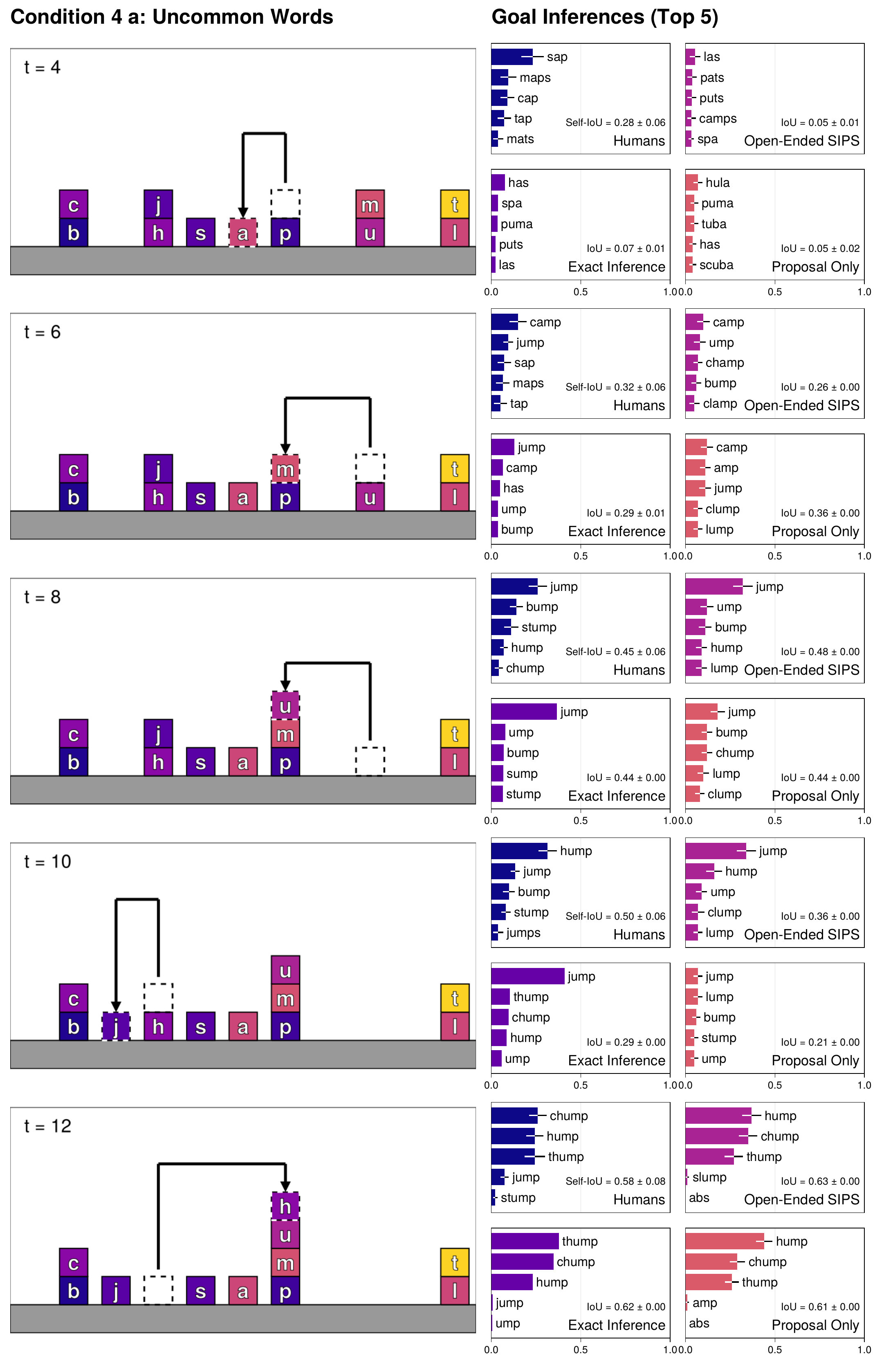}
    \caption{Uncommon Words scenario: \textbf{\textsf{chump}}}
    \end{subfigure}
    \caption{Step-by-step inference results on scenarios from the \emph{Bottom-Up Friendly} and \emph{Uncommon Words} conditions. In \textbf{(a)} there are few qualitative differences between the inference methods, apart from $t = 8$, where the proposal only baseline generates words that end in \textbf{\textsf{p}} instead of taking into account the fact that \textbf{\textsf{n}} was previously stacked on \textbf{\textsf{s}}. In \textbf{(b)}, humans, open-ended SIPS, and the bottom-up proposal largely fail to guess the uncommon word \textbf{\textsf{chump}} until the very last timestep $(t = 12)$, in contrast to the fully enumerative baseline, which assigns a non-trival probability to \textbf{\textsf{chump}} by $t = 10$.}
    \label{fig:storyboards-appendix}
\end{figure*}

In Figure \ref{fig:storyboards-appendix}, we show step-by-step comparisons between human and model inferences for the two experimental conditions not covered in the main text. 

In the \emph{Bottom-Up Friendly} scenario, limited inverse planning is necessary, and so the bottom-up proposal is as good an explanation for human goal inference as open-ended SIPS for all steps except $t=8$. At this step, the bottom-up proposal generates words that end in \textbf{\textsf{p}} by following the \textsc{last-tower} proposal strategy. This fails to take into account the previous actions of stacking block \textbf{\textsf{n}} onto block \textbf{\textsf{s}}, highlighting the importance of inverse planning in even simple scenarios.

In the \emph{Uncommon Words} scenario, we chose the uncommon word \textbf{\textsf{chump}} to be the goal, and designed actions to make more common distractors like \textbf{\textsf{jump}} and \textbf{\textsf{hump}} seem likely. As expected, most humans failed to think of \textbf{\textsf{chump}} as a possibility until the very last step $(t = 12)$. Open-ended SIPS and the proposal only baseline with $N = 2$ particles reflected this tendency, demonstrating how inferring rare events is difficult with a small number of samples. In contrast, the exact inference baseline enumerates all possible words at every step, and hence assigns \textbf{\textsf{chump}} a significant probability at $t = 10$. This is the case even though \textbf{\textsf{chump}} is less likely under the goal prior $P(g)$: an event being rare under the proposal $Q$ leads to qualitatively different behavior than exact inference
over events with low prior probabilities.

\subsection{Accuracy and Runtime}

\begin{figure}[t]
    \centering
    \begin{subfigure}[b]{\columnwidth}
    \includegraphics[width=\columnwidth]{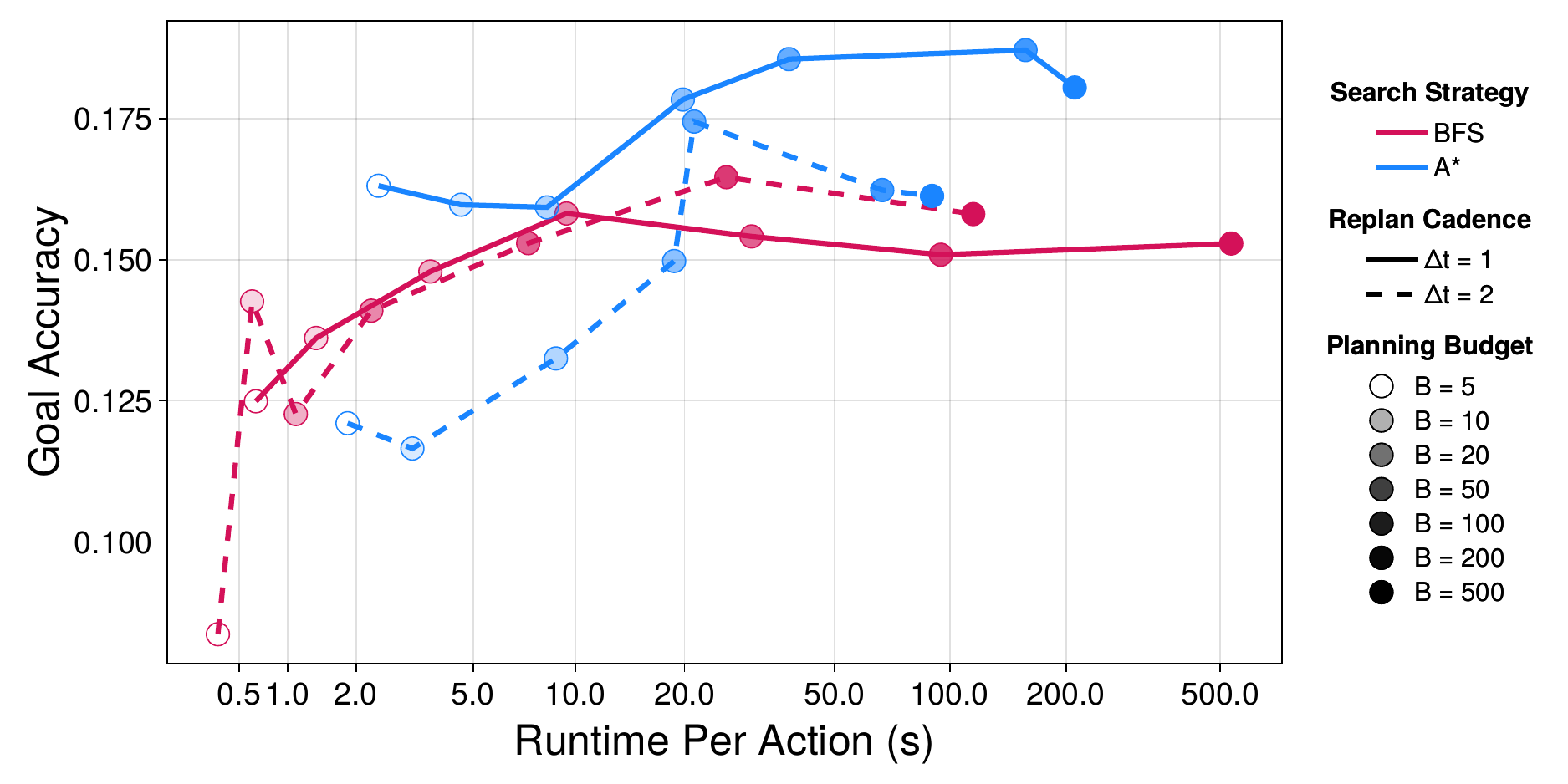}
    \caption{Exact Inference}
    \end{subfigure}
    \begin{subfigure}[b]{\columnwidth}
    \includegraphics[width=\columnwidth]{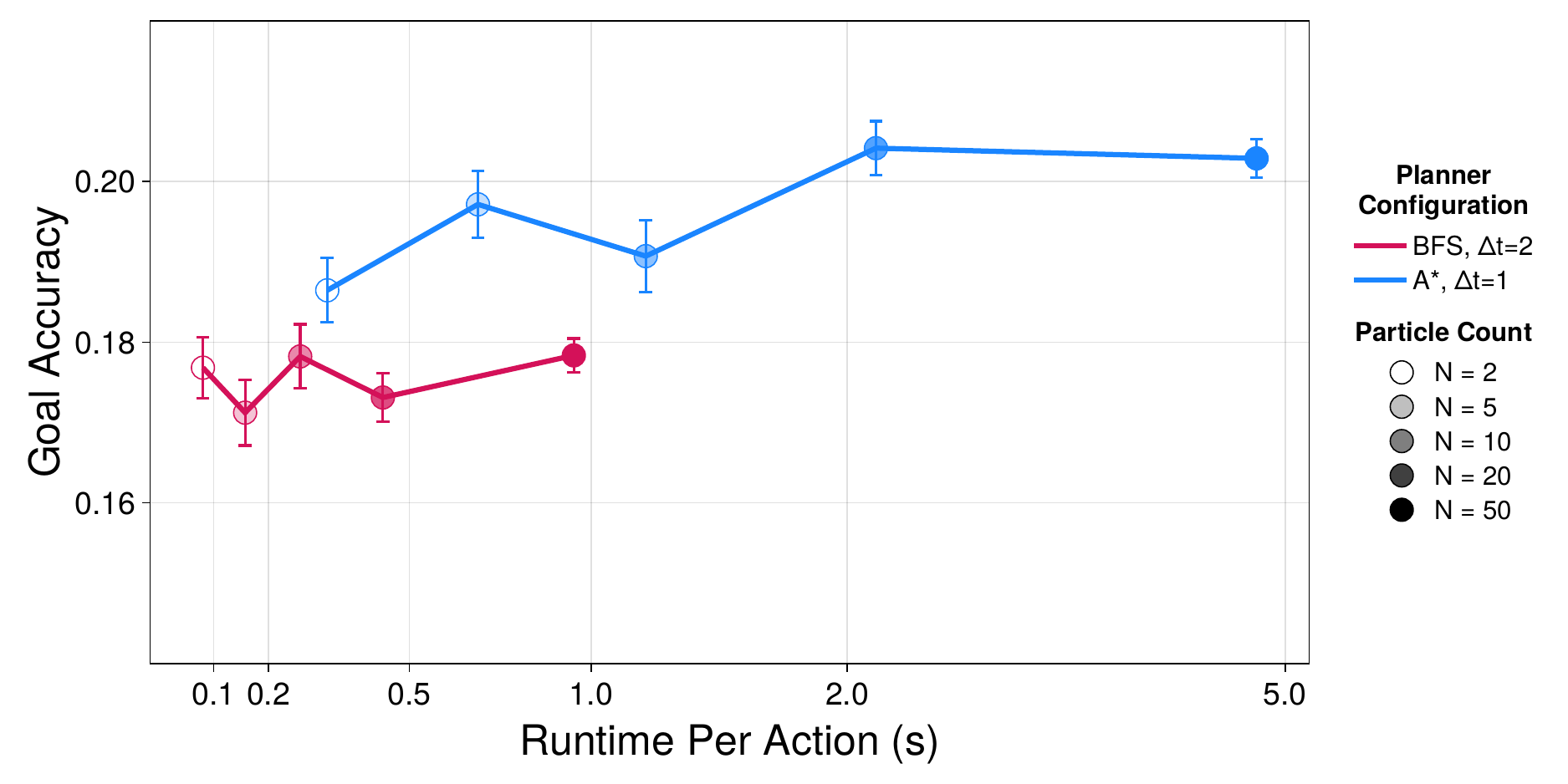}
    \caption{Open-Ended SIPS}
    \end{subfigure}
    \caption{Accuracy vs. runtime for \textbf{(a)} exact inference and \textbf{(b)} open-ended SIPS across planner configurations.}
    \label{fig:accuracy-vs-runtime}
    \vspace{-9pt}
\end{figure}

While our analysis in this paper focused on open-ended SIPS as a rational process model of human goal inference, our algorithm can also be used as a practical tool for building AI systems that better infer people's goals in order to assist them \shortcite{zhixuan2024pragmatic}. To that end, we compare the accuracy-runtime tradeoffs of both open-ended SIPS and the baseline methods. All experiments were conducted on a laptop with an i7-1370P 1.90 GHz CPU and 64 GB of RAM.

\textbf{Effect of planner configuration.} In Figure \ref{fig:accuracy-vs-runtime}, we show how the accuracy of goal inference changes as a function of the planner configuration used in the generative model. Accuracy is plotted against algorithm runtime per observed action. We find that accuracy generally increases with planning budget, indicating the importance of spending enough computation on calculating good $\hat Q_{\pi_t}$ estimates, which improves the quality of the action likelihood $P(a_t | g)$. The effect of the search strategy is more subtle. Using A* search as the RTHS search strategy can \emph{improve} accuracy when $\Delta t = 1$ (i.e. when the policy is updated at every timestep). However, this leads to an increased runtime compared to BFS for any given planning budget $B$, with A* being 2-5 times slower than BFS for low planning budgets. Still, using A* with $\Delta t = 1$ achieves higher accuracy than BFS ever does, indicating its value when accuracy is paramount. Unlike the planning budget, increasing the particle count $N$ does not appear to substantially improve the average accuracy of open-ended SIPS, with changes in planner configuration dominating any accuracy improvement from additional particles.

\textbf{Overall comparison.} In Table \ref{tab:performance}, we show accuracy and runtime measures across all inference methods and particle counts. For the inverse planning methods, we fixed the planning budget to $B = 100$, but compare both the accuracy-maximizing planner configuration ($\sigma = \text{A*}, \Delta t = 1$) and the runtime-minimizing configuration ($\sigma = \text{BFS}, \Delta t = 2$). While exact inference has zero variance, the cost of tracking all goal hypotheses leads to an unacceptably high runtime for many applications. The bottom-up proposal is on the opposite end of the spectrum, achieving millisecond or less runtimes but with a subpar accuracy that does not increase with particle count. In contrast, open-ended SIPS combines the best of both worlds, achieving the highest goal accuracies while maintaining real-time speed. Changing the particle count $N$ trades off variance in inference results vs. runtime. By using $N=20$, the variance in the probability estimate of the true goal can also be limited to less than 10\% while still taking less than half a second process each new observation. These results demonstrate the promise of open-ended SIPS as a practical algorithm for real-time open-ended goal inference.

\begin{table}[t]
\scriptsize
\centering
\begin{tabular}{@{}llrrrr@{}}
\toprule
\textbf{Method} & \textbf{\begin{tabular}[c]{@{}l@{}}Search\\ Strategy\end{tabular}} & \multicolumn{1}{l}{\textbf{\begin{tabular}[c]{@{}l@{}}Particle\\ Count $N$\end{tabular}}} & \multicolumn{1}{l}{\textbf{\begin{tabular}[c]{@{}l@{}}Runtime /\\ Act. (s)\end{tabular}}} & \multicolumn{1}{l}{\textbf{\begin{tabular}[c]{@{}l@{}}Accuracy\\ $P(g_\text{true})$\end{tabular}}} & \multicolumn{1}{l}{\textbf{\begin{tabular}[c]{@{}l@{}}Accuracy\\ Std. Dev.\end{tabular}}} \\ \midrule
Exact           & BFS                                                                & ---                                                                                       & \textbf{7.31}                                                                             & 0.153                                                                                              & \textit{\textbf{0.000}}                                                                   \\
Inference       & A*                                                                 & ---                                                                                       & 37.95                                                                                     & \textbf{0.186}                                                                            & \textit{\textbf{0.000}}                                                                   \\ \midrule
Open-Ended      & BFS                                                                & 2                                                                                         & \textbf{0.08}                                                                             & 0.177                                                                                              & 0.218                                                                                     \\
SIPS            &                                                                    & 5                                                                                         & 0.16                                                                                      & 0.171                                                                                              & 0.140                                                                                     \\
                &                                                                    & 10                                                                                        & 0.26                                                                                      & \textbf{0.178}                                                                                     & 0.095                                                                                     \\
                &                                                                    & 20                                                                                        & 0.44                                                                                      & 0.173                                                                                              & 0.069                                                                                     \\
                &                                                                    & 50                                                                                        & 0.95                                                                                      & 0.178                                                                                              & \textbf{0.049}                                                                            \\ \midrule
Open-Ended      & A*                                                                 & 2                                                                                         & \textbf{0.32}                                                                             & 0.186                                                                                              & 0.225                                                                                     \\
SIPS            &                                                                    & 5                                                                                         & 0.67                                                                                      & 0.197                                                                                              & 0.146                                                                                     \\
                &                                                                    & 10                                                                                        & 1.18                                                                                      & 0.191                                                                                              & 0.108                                                                                     \\
                &                                                                    & 20                                                                                        & 2.14                                                                                      & \textit{\textbf{0.204}}                                                                            & 0.080                                                                                     \\
                &                                                                    & 50                                                                                        & 4.73                                                                                      & 0.203                                                                                              & \textbf{0.055}                                                                            \\ \midrule
Proposal        & ---                                                                & 2                                                                                         & \textit{\textbf{0.0004}}                                                                  & 0.145                                                                                              & 0.171                                                                                     \\
Only            &                                                                    & 5                                                                                         & 0.0006                                                                                    & \textbf{0.147}                                                                                     & 0.104                                                                                     \\
                &                                                                    & 10                                                                                        & 0.0008                                                                                    & 0.139                                                                                              & 0.073                                                                                     \\
                &                                                                    & 20                                                                                        & 0.001                                                                                     & \textbf{0.147}                                                                                     & 0.052                                                                                     \\
                &                                                                    & 50                                                                                        & 0.003                                                                                     & 0.146                                                                                              & \textbf{0.035}                                                                            \\ \bottomrule
\end{tabular}
\caption{Accuracy and runtime across inference methods, RTHS search strategies, and particle counts. \textbf{Bold} entries denote the best performance within each method. \textit{Italicized} entries denote best performance across methods.}
\label{tab:performance}
\vspace{-9pt}
\end{table}

\end{document}